\documentclass{article}
\usepackage{graphicx}

\usepackage{amssymb}
\usepackage{amsmath}
\usepackage{xcolor}
\usepackage{hyperref}

\usepackage{natbib}
\newcommand{\changed}[1]{\textcolor{black}{#1}}
\usepackage{amsmath,amsfonts}
\usepackage{algorithmic}
\usepackage{algorithm}
\usepackage{array}
\usepackage{textcomp}
\usepackage{stfloats}
\usepackage{url}
\usepackage{graphicx}

\usepackage{verbatim}

\usepackage{caption}
\usepackage{subcaption}

\begin{document}



\title{Landscape Features in Single-Objective Continuous Optimization: Have We Hit a Wall in Algorithm Selection Generalization?} 

\author{Gjorgjina Cenikj, Gašper Petelin, Moritz Seiler, Nikola Cenikj, Tome Eftimov}

\maketitle

\begin{abstract}
The process of identifying the most suitable optimization algorithm for a specific problem, referred to as algorithm selection (AS), entails training models that leverage problem landscape features to forecast algorithm performance. A significant challenge in this domain is ensuring that AS models can generalize effectively to novel, unseen problems. This study evaluates the generalizability of AS models based on different problem representations in the context of single-objective continuous optimization. In particular, it considers the most widely used Exploratory Landscape Analysis features, as well as recently proposed Topological Landscape Analysis features, and features based on deep learning, such as DeepELA, TransOptAS and Doe2Vec. Our results indicate that when presented with out-of-distribution evaluation data, none of the feature-based AS models outperform a simple baseline model, i.e., a Single Best Solver.
\end{abstract}



\section{Introduction}

Motivated by the potential to capitalize on the varied performance of different algorithms across sets of different problem instances, the algorithm selection (AS) task targets the automated identification of a preferred optimization algorithm to solve a particular problem instance~\cite{as_combinatorics, as_survey}. 

Conventionally, AS is performed by taking into account the properties of the problem instance, which are typically described in the form of a numerical vector representation, also referred to as problem landscape features. Once a problem instance is represented in a vector form, Machine Learning (ML) models can be used to capture the relation between problem landscape features and algorithm performance, and further identify the best algorithm for a problem instance.

In the field of single-objective continuous optimization, the most common choice of problem landscape features used to represent problem instances are the Exploratory Landscape Analysis (ELA)~\citep{ela} features. ELA features are useful for characterizing optimization problems, but they also have some limitations, such as being sensitive to the sample size and the sampling method used to sample solutions from the decision space~\citep{ela_sensitive_to_sampling, urban_ela_not_invariant_sampling}, and not being invariant to transformations such as scaling and shifting of the problem~\citep{urban_ela_not_invariant_sampling, urban_ela_not_invariant}. Furthermore, AS models using ELA features have shown limited generalizability to unseen problems~\citep{urban_transfer_learning, gasper_affine_generalization,gina_cross_benchmark_generalization}.

To address this issue, several research works in the past few years have proposed the development of alternative representations of continuous problem instances~\citep{doe2vec, transopt, tla}, however, as of yet, their utility for the AS task and the degree to which they improve the generalizability of an AS model has not been evaluated.

In the field of single-objective continuous optimization, AS is primarily conducted using the established Black-Box Optimization Benchmarking (BBOB)~\citep{bbob} suite, containing 24 classes of optimization problems. Different instances can be generated from each problem class by applying transformations such as scaling or shifting to the original problems. Such a generation process makes the problem instances within the same problem class highly similar to each other. When using the BBOB benchmark for AS and performance prediction, there are substantial differences in how model evaluation is performed. The two commonly used approaches are referred to as ``leave-instance-out" (LIO) and ``leave-problem-out" (LPO). LIO is the easiest evaluation strategy where instances from all problem classes are present in the training set, but only a few instances from each class are used for testing. This means the AS model encounters functions with similar problem landscape features during training and testing. On the other hand, the LPO validation strategy involves testing the model on problem classes which were not present in the training data. Due to the diversity of problems in the BBOB benchmark, the LPO evaluation strategy is much more challanging, since the test set may contain problem instances which are not similar to any of the instances seen during training. 
Due to such discrepancies in the difficulty of the evaluation, choosing which strategy to use can substantially influence the results~\citep{tanabe2022benchmarking}.
To mitigate this issue and improve the diversity of problems provided in the BBOB benchmark, affine combinations of the original BBOB problem instances have been generated~\citep{affine_problems, affine_diederick}. 

\textbf{Our contribution:} In this work, we leverage a benchmark composed of affine combinations of BBOB problem instance to analyze the generalizability of AS models using state-of-the-art problem landscape features.
We aim to provide a comprehensive analysis of recently proposed problem landscape features based on deep learning and topological landscape analysis, and compare them to the most commonly used ELA feature groups. In particular, we focus on the TinyTLA~\citep{tinytla}, Doe2Vec~\citep{doe2vec}, TransOptAS~\citep{transoptas}, and the DeepELA~\citep{deepela} features.
We selected these feature groups due to their recent introduction (all of them have been developed in the last two years), and the fact that there has been no comparative analysis conducted among all of them thus far. Furthermore, these features do not require additional function evaluations and can be calculated on the same sample of problem solutions, enabling a fair feature comparison. 

The main objectives of the study are the following:

\textit{i)} Analyzing the complementarity of the feature groups

\textit{ii)} Analyzing the extent to which each feature group captures algorithm performance

\textit{iii)} Evaluating the generalizability of the AS models obtained with each feature group and combinations of them. The features are evaluated for the AS task on the affine problems~\citep{affine_diederick}, using different strategies to split the problem instances into training and evaluation splits, in order to understand the degree of generalizability provided by each feature group.

\textbf{Outline:} The remainder of the paper is structured as follows. Section~\ref{section_related_work} provides some background on the investigated feature groups and related studies investigating the generalizability of AS models. Section~\ref{section_methodology} describes the data and methodology. Section~\ref{section_results} contains the results, addressing each of our objectives. Section~\ref{section_discussion} discusses our findings, while Section~\ref{section_conclusion} concludes the paper.

\section{Related work}
\label{section_related_work}
In this section, we introduce the investigated feature groups and present related works on the topic of generalizability of AS models in single-objective continuous optimization.

\subsection{Feature Representations}

A prerequisite for training a ML model to AS is to first represent optimization problems with a set of numerical features. 
All of the investigated features (except for a few of the ELA feature groups) are calculated on a ``static" sample of solutions which are sampled from the problem instance using a predefined sampling strategy such as Latin Hypercube Sampling~\citep{lhs} or random sampling. Therefore, they generate a global representation of the problem instance, regardless of the algorithm's behaviour.
There are also features which are extracted from the solutions sampled by the optimization algorithms while they are being executed on a given problem instance. Some examples include the DynamoRep~\citep{cenikj2023dynamorep}, Opt2Vec~\citep{opt2vec}, and features based on probing trajectories~\citep{probing_trajectories}. However, the focus of our study is on features which are based on a ``static" sample of problem solutions, whose computation does not require additional function evaluations or algorithm execution. While we consider the trajectory-based features to have great potential, their comparison is beyond the scope of this work, since a large number of features is already obtained using the approaches being considered.
In this remainder of this subsection, we present the investigated feature groups in more detail.
\subsubsection{ELA}
Exploratory Landscape Analysis (ELA)~\citep{ela} is an approach to characterize black-box optimization problem instances using a set of mathematical and statistical low-level features. They aim to capture various aspects of the problem instance, such as its curvature, convexity, properties of the distribution of objective function values (its skewness, kurtosis, or number of peaks), the multimodality, sizes of basins of attraction, etc.
ELA features can be further distinguished into cheap features that are computed from a fixed set of samples and expensive features that require additional sampling.

\subsubsection{TinyTLA}
Topological Landscape Analysis (TLA)~\citep{tla} features identify landscape features of optimization problems using Topological Data Analysis (TDA). TDA involves analyzing a set of data points, often called a point cloud, to discover topological structures such as spheres, torus, connected components, or even more complicated surfaces and manifolds. Identifying these structures allows for the description of the point cloud through features that reflect the presence of these structures at varying scales, independently of certain transformations (e.g. rotations, scaling). 
In addition to the original TLA, a refined version named TinyTLA~\citep{tinytla} has been introduced, which calculates features with fewer data samples. The TinyTLA features have been evaluated in the BBOB problem classification context, with a thorough examination of how parameters affect them and a preliminary assessment for the AS task within the BBOB benchmark.

\subsubsection{Doe2Vec}
Doe2Vec~\citep{doe2vec} is a deep-learning-based methodology that uses a variational autoencoder to learn optimization landscape representations in an unsupervised way. Given a sample of objective function values from the problem instance, the autoencoder is trained to reconstruct the sample, and in this way, it learns an informative latent representation of the problem. The Doe2Vec model is an encoder-decoder structure with several fully-connected layers. It is trained on a large set of problem instances generated with a random function generator~\citep{random_function_generator_python}. The Doe2Vec features have demonstrated to be promising for identifying similar surrogate problem instances that are inexpensive to evaluate. They have also been shown to enhance performance when used alongside traditional ELA features for the task of recognizing high-level properties of BBOB problem instances. To the best of our knowledge, this is the first study which evaluates the Doe2Vec features for the AS task. 

\subsubsection{TransOptAS}
The TransOptAS features~\citep{transoptas} are derived from a transformer-based model trained to predict the performance of various optimization algorithms. The model receives as input raw candidate solutions and their corresponding objective values. The encoder of the transformer model processes these samples to create representations for each candidate solution. The mean of the produced representations is used to represent the entire sample. This representation is further passed to a regression head, which outputs a numerical performance indicator in the range [0,1] for each algorithm in the portfolio.
The model is trained on a set of random functions generated by the same generator used in the Doe2Vec methodology.
\subsubsection{DeepELA}
DeepELA~\citep{deepela} is another methodology based on a transformer architecture. Unlike the TransOptAS model, which is trained in a supervised manner, the DeepELA methodology uses a self-supervised training of the transformer architecture to produce problem representations which are invariant to problem transformations.
This is achieved by training the model to produce the same representation for a problem instance and a variant of it which has undergone a transformation which does not change its underlying properties (rotation, inversion of decision variables or randomization of the sequence of decision and objective variables). Two studies evaluating the complementarity of DeepELA and classical ELA features were recently presented in~\citep{deep_classical_LION,seiler2024LearnedVsClassic}.

\subsection{Analyses of Algorithm Selection Generalizability}

The ultimate goal of performing AS is to train a ML model that can recommend the most suitable optimization algorithm for completely new optimization problems with different landscape features. Unfortunately, the generalization of AS models to completely new problems is not easily achieved.

In~\citep{urban_transfer_learning}, it was shown that an AS model using ELA features, trained on randomly generated functions~\citep{random_function_generator_matlab} exhibits poor generalization to the BBOB benchmark. This observation has been confirmed across multiple studies conducted on different benchmarks.

The generalization of AS models across four different benchmarks has been explored in~\citep{gina_cross_benchmark_generalization}. In this case, the experimental setup involves the use of an entire benchmark is used to train the AS model, and it is evaluated on a completely different benchmark. This work also investigates the complimentarity between the ELA and transformer features trained for the task of BBOB problem classification~\citep{transopt}. It is observed that when the training and testing benchmarks have a similar distribution of algorithm performance and share the same single-best solver, the feature-based AS models cannot outperform a simple baseline model predicting the mean performance on the training benchmark, i.e, the single best solver. The study also points out that there are problem instances for which there is no guarantee that similar problem features will lead to similar algorithm performance, meaning that the investigated features lack the discriminatory power to accurately capture algorithm performance.

The generalization of an AS model using ELA features trained on the original BBOB problem instances and evaluated on their affine combinations has been explored in~\citep{gasper_affine_generalization}. The results suggest that while the AS predicts the ranks of algorithms for problem instances similar to the ones observed during training relatively well, its performance substantially declines as the problems instances used for evaluation deviate from the training distribution. In these cases, the model's predictive performance becomes comparable to a baseline model predicting the mean rank of the algorithms in the training set. Our study performs a similar analysis, with the difference that ~\citep{gasper_affine_generalization} is focused on analyzing ELA feature groups, while ours also targets the comparison of recently proposed features based on deep learning and TDA.

In~\citep{assessing_generalizability}, a methodology for assessing how well a predictive model, developed for algorithm performance using one benchmark, can be applied to another. The findings indicate that the patterns of generalizability observed within the problem landscape features are also mirrored in the algorithm performance outcomes.

While most of these studies analyze the generalizability of only the ELA features, (with the exception of the inclusion of transformer features in ~\citep{gina_cross_benchmark_generalization}), our study aims to provide a more comprehensive view of recently proposed features whose generalizability is yet to be investigated.

\section{Methodology}
\label{section_methodology}
In this section, we describe the methodology, starting off by introducing the problem and algorithm portfolios, followed by the description of the metric used for algorithm performance evaluation. We then describe the feature calculation procedure, the training of the ML model used for AS, as well as the different settings used for its evaluation.
\subsection{Problem Portfolio}
We use the first five problem instances from the 24 BBOB problem classes to generate affine recombinations as proposed in~\citep{affine_diederick}. The affine transformations have been generated using the equation suggested in~\citep{affine_diederick}, also presented below:

\begin{equation}
    \begin{matrix}
F(P_{i, m}, P_{j, n}, \alpha)(x) = \qquad \qquad  \qquad \qquad  \qquad  \qquad   \\ 
 \exp(\alpha \, log(P_{i, m}(x) - P_{i, m}(O_{i, m})) + \qquad \qquad \\ 
 \qquad (1-\alpha) \, log(P_{j, n}(x-O_{i, m}+O_{j, n}) - P_{j, n}(O_{j, n}))).
\end{matrix}
\end{equation}
In this case, $P_{a, b}$ denotes the $b$-th instance of the problem within the $a$-th BBOB problem class. Meanwhile, $O_{a, b}$ represents the location of the optimum for the function $P_{a, b}$. The parameter $\alpha$ illustrates the blending extent of the two functions. 

In particular, we use the first five instances of the all of the 24 BBOB problem classes, $m,n \in \{1,\ldots,5\}$. We combine problem instances from different problem classes only if they have the same instance identifier, i.e., $ i \neq j, m = n$. For example, the first instance of the first problem class is combined with the first instances of the remaining 23 problem classes. The recombination is performed with $\alpha$ values of 0.25, 0.50, and 0.75 for all pairs of problem instances, obtaining a total of 8,280 problem instances.

We use Latin Hypercube Sampling~\citep{lhs} to construct a sample of size 50$d$ ($d$ being the problem dimension), on which we evaluate all of the problem instances to obtain their objective function values.

\subsection{Algorithm Portfolios}
We use 4 algorithm portfolios, which contain different configurations of the Differential Evolution (DE)~\citep{de} and Particle Swarm Optimization (PSO)~\citep{pso} algorithms. For the DE algorithm, we are randomly initializing the ``CR" (crossover constant), ``F" (weighting factor), and ``variant" (controling the crossover and selection strategies) parameters.

For the PSO algorithm, we are configuring the ``initial velocity" parameter (which takes values of ``random" when the velocity vector is initialized randomly or ``zero" when the particles start to find the direction through the velocity update equation), the ``adaptive" parameter which controls whether the inertia, social and coginitive impact are changed dynamically or not, and ``w" indicating the inertia value.

The four different algorithm portfolios that we evaluate are as follows:

\begin{itemize}
    \item The 5DE algorithm portfolio contains 5 different configurations of the DE algorithm, whose parameter initialization can be found in Table~\ref{tab:de_configs}.
    \item The 5PSO algorithm portfolio contains 5 different configurations of the PSO algorithm, whose parameter initialization can be found in Table~\ref{tab:pso_configs}.
    \item The 5DE+5PSO algorithm portfolio contains both the DE and PSO algorithms used in portfolios 5DE and 5PSO.
    \item The 2DE+2PSO algorithm portfolio contains the 2 best performing algorithms from each of the 5DE and 5PSO algorithms portfolios
\end{itemize}

We use multiple algorithm portfolios in order to investigate both the scenarios where all of the algorithms come from the same algorithm class (automatic algorithm configuration), and the scenarios where the algorithms can belong to different algorithm classes (automatic algorithm selection). Using multiple algorithm portfolio also enabels us to analyze the performance of the TransOpt features trained for one algorithm portfolio, and applied on another algorithm portfolio.

\begin{table}[h!]
\centering
\caption{Parameters of the DE configurations}
\begin{tabular}{|c|c|c|c|c|}

\hline
& CR & F & selection & crossover \\ \hline
DE1 & 0.5 & 0.5 & random & binary \\ \hline
DE2 & 0.6 & 0.4 & best & exponential \\ \hline
DE3 & 0.9 & 0.7 & best & exponential \\ \hline
DE4 & 0.3 & 0.8 & best & binary \\ \hline
DE5 & 0.8 & 0.4 & best & binary \\ \hline
\end{tabular}
\label{tab:de_configs}
\end{table}

\begin{table}[h!]
\centering
\caption{Parameters of the PSO configurations}
\begin{tabular}{|c|c|c|c|}
\hline
& initial velocity & adaptive & weight \\ \hline
PSO1 & zero & True & 0.9 \\ \hline
PSO2 & zero & False & 0.9 \\ \hline
PSO3 & random & True & 0.9 \\ \hline
PSO4 & random & False & 0.9 \\ \hline
PSO5 & random & False & 0.7 \\ \hline
\end{tabular}
\label{tab:pso_configs}
\end{table}

\subsection{Algorithm Performance Metric}
\label{subsec:algorithm_performance_metric}

We use a fixed-budget scenario to assess the algorithms, where the evaluation involves recording the best-found objective function value after a specified budget of algorithm iterations. The algorithm performance is captured using the normalized precision metric introduced in~\citep{gina_cross_benchmark_generalization} and is defined as follows.

Let $y_{arp}$ denote the objective function value of the best solution found by algorithm $a$ in run $r$ for problem instance $p$. This value, $y_{arp}$, is normalized based on the range of solutions found by the remaining algorithms for the same problem instance and initial population (same run). Representing by $b_{rp}$ and $w_{rp}$ the best and worst solutions found for problem instance $p$ in run $r$ by any of the algorithms, we define the scaled best objective function value achieved by algorithm $a$ in run $r$ on problem instance $p$:

   \begin{equation}
        s_{arp} = \frac{y_{arp} - b_{rp}}{w_{rp} - b_{rp}}.
        \label{eq:algorithm_performance_metric}
    \end{equation}
This metric quantifies the extent to which the best solution found by algorithm $a$ in run $r$ outperforms other algorithms executed with the same initial population. The normalized precision score of algorithm $a$ on problem instance $p$ across all runs is then calculated by taking the mean of all $s_{arp}$.

\subsection{Feature Calculation}
In this subsection, we explain the calculation of each of the investigated feature groups.
\subsubsection{ELA}

 We compute ELA features from the following feature groups using the \texttt{flacco} Python library~\citep{pflacco}:  \texttt{disp}, \texttt{ela\_distr}, \texttt{ela\_level}, \texttt{ela\_meta}, \texttt{ic}, \texttt{nbc}, and \texttt{pca}. In this way, we obtain a total of 62 features for each problem instance. Please note that we selected these features groups since they do not require additional function evaluations (which is the case for other feature groups such as \texttt{local\_search}, \texttt{convexity} and \texttt{curvature} features), and can hence be calculated on the same sample of solutions as the other feature groups for a fair comparison.
In~\citep{ela_y_normalization, gina_ela_generalization_scaling}, it has been shown that calculating ELA features on a sample where the objective function values scaled to be in the range [0,1] using min-max scaling can improve their generalizability. The methodology of the other investigated feature groups also includes some type of scaling of the objective function values. For these reasons, we additionally evaluate a version of the ELA features calculated on samples where the objective function values are scaled in the range [0,1] using min-max scaling. We refer to these features as \texttt{ela\_scaled}.

\subsubsection{TinyTLA}

In the process of topological feature extraction, problem samples are treated as a point cloud from which features are extracted. Both the candidate solutions (x-values) and objective function values (y-values) are first scaled to be between 0 and 1 for each problem instance separately. A \textit{volume} transformation is applied to the x-values as described in~\citep{tinytla}. The goal of this transformation is to reduce the features' sensitivity to some problem transformations such as shifting and scaling.
The euclidean distances between the x-values and y-values for each pair of samples are calculated independently. The final distances between each pair of samples are obtained by calculating the weighted sum of the distances between the x-values and the y-values. In our case, the weight parameter for combining the distances, $\alpha$ is set to 0.3. 

The thereby obtained distance matrix is used to compute persistence diagrams, which encapsulate information about topological structures in the point cloud. Subsequently, numerical features are constructed by transforming the persistence diagram into a persistence image, whose pixels produce a high-dimensional vector that characterizes the problem instance in terms of topological properties.
The size of the kernel used to map the persistence diagrams into persistence images is set to 0.002. 

The selection of all parameter values was done with consultation of the TinyTLA creators, and based on the analysis of the best-performing parameter combinations investigated in \citep{tinytla}.

The above feature calculation procedure is repeated for homologies of dimensions 0, 1, and 2 ($H_0$, $H_1$ and $H_2$), which capture different types of topological structures. The 0-dimensional homology $H_0$ describes the connected components of a simplicial complex, the 1-dimensional homology $H_1$ represents 1-dimensional loops (structured like a doughnut), while the 2-dimensional homology $H_2$ captures 2-dimensional cavities. We obtain 50 features from $H_0$ and 2,500 features from each of $H_1$ and $H_2$. The features from the three homologies are concatenated into a single vector, so we obtain a total of 5,050 features from the TinyTLA approach.
For more details on this methodology, we refer the reader to the original article~\citep{tinytla}.

\subsubsection{Doe2Vec}
The training of the Doe2Vec model follows the default implementation in the \texttt{doe2vec} python library~\citep{doe2vec_python}. The objective function values are scaled in the range [0,1]. The variational autoencoder is trained on a set of 250,000 randomly generated functions for a total of 100 epochs. The dimension of the latent representation is set to 32, meaning that the model produces a representation of size 32 for each problem instance.
We use a custom sample of candidate solutions to evaluate the problems generated by the random function generator, obtain their objective function values, and train the model. The same sample is then used to evaluate the affine problem instances, obtain their objective function values, and pass them through the model to obtain their latent representation used as problem landscape features. This is done so that the sample is fixed across all feature groups to ensure a fair comparison.

\subsubsection{TransOptAS}
The TransOptAS model is trained to predict algorithm performance on a set of 30,000 randomly generated problems. All algorithms are executed on these functions to obtain the performance data. The model includes two layers in the encoder of the transformer architecture, each one consisting of two heads. The size of the embedding is set to 50, meaning that the generated problem representations are of size 50. The model is trained for 100 epochs using an early stopping strategy which terminates the training early if an improvement of 0.001 in the validation loss is observed for 3 epochs. 
Once the model is trained, the weights are frozen and the regression layer (which predicts algorithm performance) is removed from the TransOptAS model. The next-to-last layer is then used for the purpose of generating problem representations. The objective function values of the problem samples are scaled in the range [0,1] and are then passed through the model to produce their representations. A separate model is trained for predicting the performance of each algorithm portfolio, resulting in the training of four models and thereby four variations of the TransOptAS features generated for each problem instance. We name these features according the the algorithm portfolio used for the model training, i.e., \texttt{transoptas\_5DE}, \texttt{transoptas\_5PSO}, \texttt{transoptas\_5DE+5PSO} and \texttt{transoptas\_2DE+2PSO}.

\subsubsection{DeepELA}
As DeepELA is already pre-trained, no further training or fine-tuning was done. Instead, the trained \texttt{Large\_50d} model (as proposed in~\citep{deepela}) was applied to the same samples of candidate solutions used for the calculation of the other features. The predicted features are guaranteed to be invariant to the order of the candidates but they are not guaranteed to be in variant to the order of the dimensions (see~\citep{seiler2024LearnedVsClassic}). Hence, we compare a single forward pass to ten repeating forward passes as outlined in~\citep{seiler2024LearnedVsClassic}. The authors introduced the latter approach to improve the robustness of the generated features but reported only a slight improvement.
Specifically, instead of performing a single forward pass to compute the feature vectors, this approach involves computing ten individual forward passes with different random permutations of the decision space to compensate for small fluctuations of the feature vectors. Afterwards, the arithmetic mean of the ten feature vectors is used as a representation. We will refer to the two approaches \texttt{deep\_ela\_large\_r1} (calculated with a single forward pass) and \texttt{deep\_ela\_large\_r10} (calculated as the mean of ten forward passes).


\subsection{Automated Algorithm Performance Prediction Model}
We use a Random Forest (RF) model to perform multi-target regression, where the model predicts the performance score of each algorithm as defined in Equation~\ref{eq:algorithm_performance_metric}.

We address the task as a multi-target regression for the following reasons:

\textit{i)} Unlike multi-class classification, which predicts a single best algorithm (as done in ~\citep{urban_transfer_learning}), this method captures the performance of all algorithms. This approach avoids penalizing a model for selecting one of two algorithms with very similar performance.

\textit{ii)} In contrast to predicting algorithm ranking (as done in ~\citep{gasper_affine_generalization}), where each algorithm is assigned an integer score based on its performance, this approach captures fine-grained differences in performance, allowing algorithms with similar best-obtained values to receive comparable scores.

We chose the RF model because it has good performance on tabular data~\citep{shwartz2022tabular}, it is robust against differently scaled feature ranges, and it provides feature importance values that help with interpretability. The RF model is executed using the default configuration parameters in the \texttt{scikit-learn}~\citep{scikit-learn} Python library version 1.2.2, meaning that the number of estimators (\texttt{n\_estimators} parameter) is 100, the criterion for measuring the quality of a split (\texttt{criterion} parameter) is the gini impurity, minimum number of samples required to split an internal node (\texttt{min\_samples\_split} parameter) is set to 2, the minimum number of samples required to be at a leaf node (\texttt{min\_samples\_leaf} parameter) is set to 1, the minimum node impurity threshold required for splitting each node (\texttt{min\_impurity\_decrease} parameter) is set to 0, the number of features to consider when looking for the best split (\texttt{max\_features} parameter) is set to the square root of the total number of features, while the rest of the parameters are set to None.

We do not perform parameter tuning in order to evaluate the impact of the proposed features only, on a fixed model configuration. 
We train a separate RF model for each algorithm portfolio and for each feature representation. Apart from considering the individual use of a single feature group, we also consider the representations obtained by concatenating pairs of feature groups, as well as concatenating all feature groups. We compare the performance of the RF models to a simple baseline ``dummy" model which predicts the mean performance of the algorithms on the train set.

\subsection{Algorithm Selector Performance Metric}

We capture the performance of the AS model by taking into consideration the true score of the algorithm predicted to be the best by the ML model:
   
       \begin{equation}
        AS\_performance = \frac{1}{ \left | \mathcal{P} \right |}  \sum_{p \in {P}} 1 - (s_{sp} - s_{bp})
        \label{eq:AS_metric}
    \end{equation}

In this case, $s_{sp}$ represents the score of the algorithm selected by the algorithm selector for problem instance $p$, which was predicted to have the best performance, while $s_{bp}$ is the score of the actual best-performing algorithm for problem instance $p$. 
The AS performance value is bounded between 0 and 1. A value of 0 occurs if the worst-performing algorithm is consistently selected as the best for all problem instances, whereas a value of 1 is attained when the algorithm selector perfectly predicts the best algorithm for every problem instance.

\subsection{Evaluation Settings}
To account for the different AS evaluation approaches present in the literature and provide a comprehensive, all-encompassing view of the contribution of the different feature groups, we explore several evaluation settings, differing in how the problem instances are split into training and testing sets.
The evaluation settings are listed in increasing level of difficulty.

\subsubsection{Instance Split}
This evaluation setting corresponds to the commonly used LIO evaluation on the BBOB benchmark suite, meaning that one instance of each generated affine problems is in the test set, while the remaining four instances are in the training set. This is the most lenient evaluation setting, since it is guaranteed that for each testing instance, the training set contains problem instances which are derived from the same pair of parent problem classes used for generating the test instance. For a specific train/test fold, the first problem instance created by combining pairs of problem classes would be placed in the test set, while the remaining instances (2-5) would be placed in the training set.

\subsubsection{Random Split}
In the random split evaluation setting, the affine problems are split into train and test set randomly, disregarding any information about the problem class or instance identifier. To provide an example, let us assume we have constructed a single train/test fold from all the affine functions. If we focus exclusively on affine pairs from problem classes 1 and 2, with five problem instances, in some folds, all five instances might appear in both the train and test sets, while in other cases, all 5 might be assigned either to the train or the test set. In other words, there is no guarantee that at least one instance of a specific pair will be in the train or test set.

\subsubsection{Problem Combination Split}
In this evaluation setting, the testing set contains all instances generated by combining one BBOB problem class with all of the remaining 23 BBOB problem classes. However, the training set does not contain any instances derived from the first problem class, and instead contains only combinations of the remaining BBOB problem classes. This means that the training set contains instances which are derived from one of the parent classes in the test set, but not the other one. For example, in one fold, in the test set we could have all problem instances which are generated by combining instances of problem class 1 with all other 23 problem classes. In this case, the training set would contain only instances obtained by combining problem classes 2-24, but would not contain any instances which use problem class 1. 

\subsubsection{Problem Split}
In this evaluation setting, the testing set contains combinations of two problem classes, while the training set contains the combinations of the remaining 22 problem classes. This is the most difficult evaluation setting, where the problem instances in the test set are guaranteed not to be derived from any problem class used in the generation of the training set.
In this case, the test set in one fold could contain all problem instances generated by combining problem class 1 only with problem class 2, while the test set would not contain any problem instance generated by combining problem class 1 with any other problem class, nor problem class 2 with any other problem class.

\section{Results}
\label{section_results}
In this section, we present the findings of the analysis of the feature groups. Our results are structured as follows:

\textit{i)} Subsection~\ref{subsec:results_algorithm_performance} provides an overview of the performance of the four algorithm portfolios.

\textit{ii)} Subsection~\ref{subsec:results_feature_complementarity} analyzes the complementarity of the investigated feature groups.

\textit{iii)} Subsection~\ref{subsec:results_feature_performance_alignment} investigates the degree to which the feature representations capture algorithm performance.

\textit{iv)} Subsection~\ref{subsec:results_AS} presents the results of the AS models using different feature types and evaluation settings.

\subsection{Algorithm Performance}
\label{subsec:results_algorithm_performance}
\begin{figure}
\begin{subfigure}{\textwidth}
\centering
\includegraphics[width=0.5\linewidth]{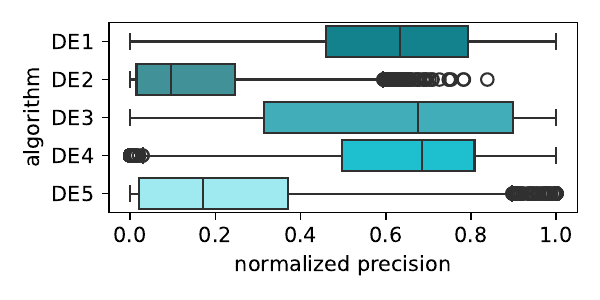}
\caption{5DE portfolio}
\label{fig:algorithm_performance_5DE}
\end{subfigure}

\begin{subfigure}{\textwidth}
\centering
\includegraphics[width=0.5\linewidth]{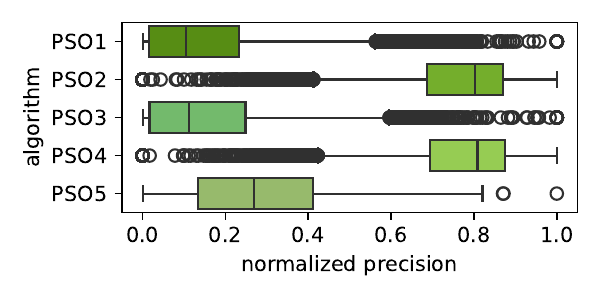}
\caption{5PSO portfolio}
\label{fig:algorithm_performance_5PSO}
\end{subfigure}

\begin{subfigure}{\textwidth}
\centering
\includegraphics[width=0.5\linewidth]{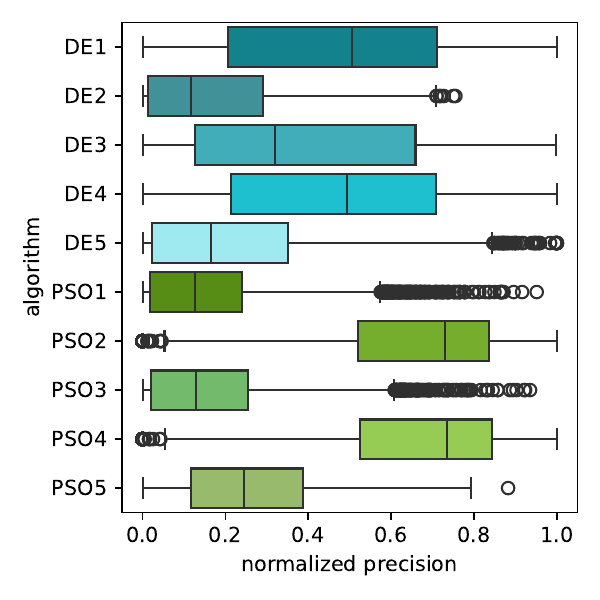}
\caption{5DE+5PSO portfolio}
\label{fig:algorithm_performance_5DE+5PSO}
\end{subfigure}

\begin{subfigure}{\textwidth}
\centering
\includegraphics[width=0.5\linewidth]{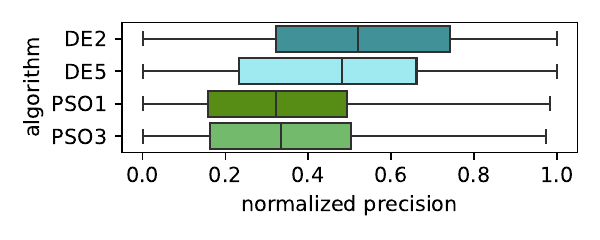}
\caption{2DE+2PSO portfolio}
\label{fig:algorithm_performance_2DE+2PSO}
\end{subfigure}
\caption{Distribution of the normalized precision scores in each algorithm portfolio. A lower normalized precision signifies a better performance of an algorithm inside a portfolio}
\label{fig:algorithm_performance}
\end{figure}

Figure~\ref{fig:algorithm_performance} shows the distribution of the normalized precision scores (defined in Equation~\ref{eq:algorithm_performance_metric}) achieved by each algorithm in the four algorithm portfolios. In the 5DE portfolio presented in Subfigure~\ref{fig:algorithm_performance_5DE}, we can see that the DE2 and DE5 algorithms provide the best results for the majority of the functions, with a median normalized precision of 0.09, and 0.17, respectively. On the other hand, the DE1, DE3 and DE4 provide median normalized precision in the range 0.63-0.68. Looking at the 5PSO portfolio in Subfigure~\ref{fig:algorithm_performance_5PSO}, we can see that the PSO1 and PSO3 algorithms, where the inertia and cognitive and social impact are changing dynamically over time, provide the best results.
In Subfigure~\ref{fig:algorithm_performance_5DE+5PSO}, presenting the 5DE+5PSO portfolio, we can see that the worst performing algorithms are PSO2 and PSO4 with a median normalized precision of around 0.72, followed by the DE1 and DE4 algorithms with a median normalized precision of approximately 0.50. The DE2, PSO1 and PSO3 have a similar median performance of around 0.13, closely followed by DE5.
Combining these algorithms in the 2DE+2PSO portfolio gives a more complementary algorithm portfolio, where the median performance of the best-performing PSO1 algorithm is around 0.33.

\subsection{Feature Complementarity}
\label{subsec:results_feature_complementarity}
\begin{figure*}
    \centering
    \includegraphics[width=\linewidth]{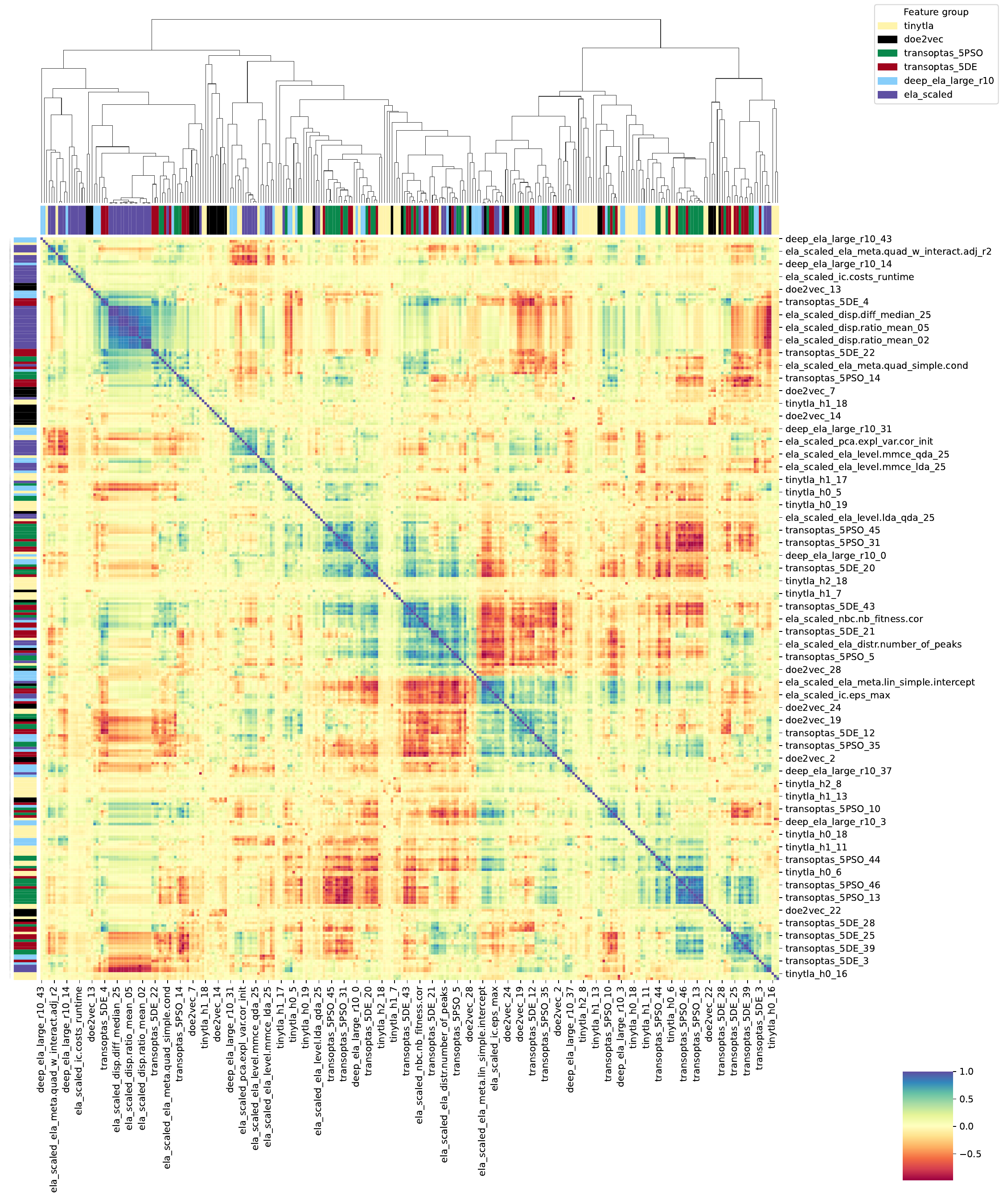}
    \caption{Spearman correlation clustermap between pairs of features from different feature groups}
    \label{fig:correlation}
\end{figure*}

Here, we investigate the complementarity of the feature groups. Figure~\ref{fig:correlation} shows a clustermap of the Spearman correlations of individual pairs of features. The Spearman correlation is calculated for each pair of features, by taking their values across all problem instances created using affine recombinations. The labels on the x and y axis indicate the names of the features being compared, while the heatmap indicates their Spearman correlation. The dendrogram is based on a hierarchical clustering performed on top of the Spearman correlations, in order to group features with similar correlations together.
Due to the high number of features, it is not possible for us to show all of the feature names in a readable way. For this reason, the feature group is additionally indicated by a color bar on the top and left side of the heatmap.

\begin{figure*}
    \centering
    \includegraphics[width=\linewidth]{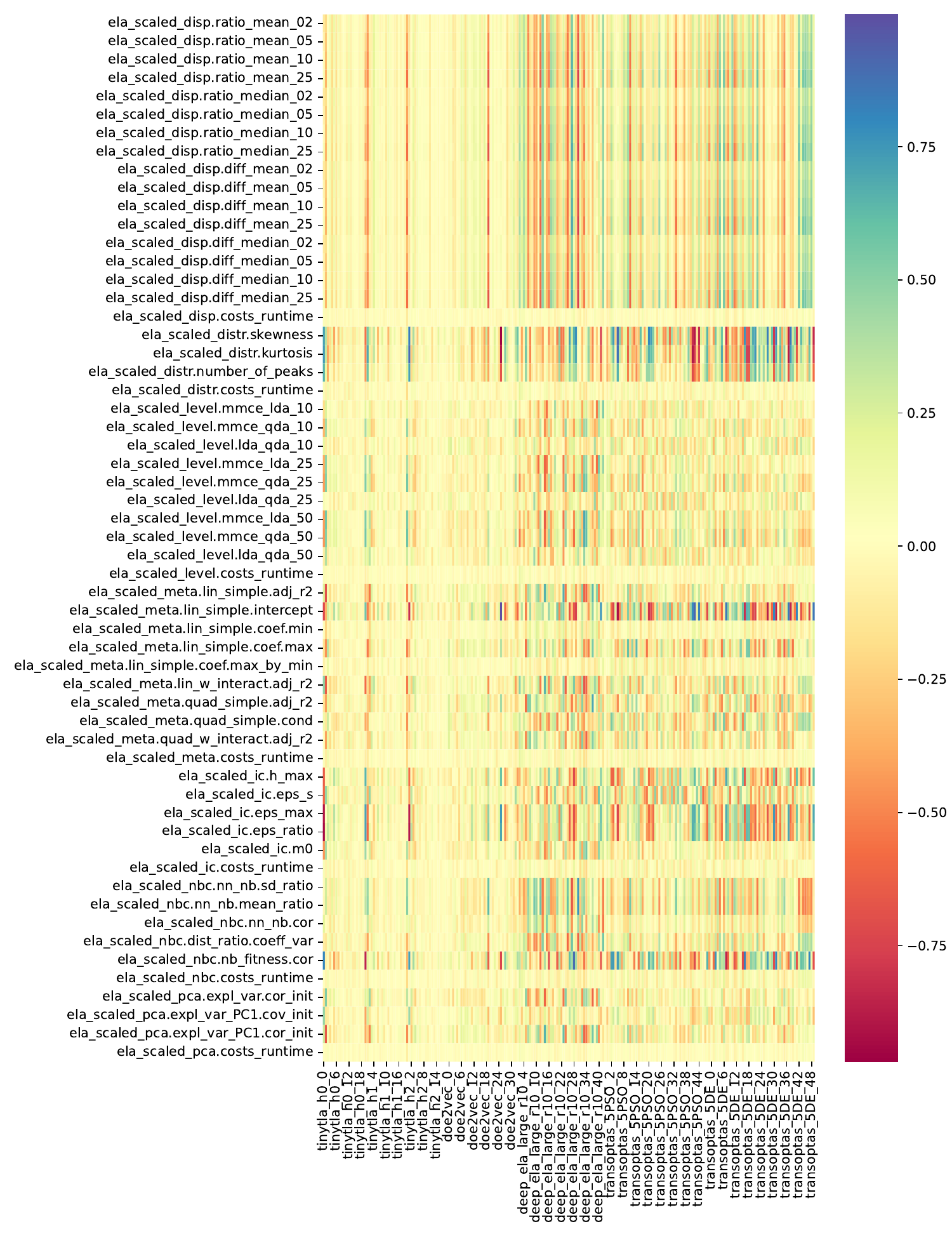}
    \caption{Spearman correlation clustermap between the \texttt{ela\_scaled} features and the other feature groups}
    \label{fig:correlation_ela_subset}
\end{figure*}

For improved visibility, we exclude some of the feature groups from this figure. In particular, we exclude the \texttt{ela} features in favor of  \texttt{ela\_scaled}, since a lot of them are not sensitive to the scaling of objective function values and are therefore identical to the \texttt{ela\_scaled} features. Similarly, we exclude the \texttt{deep\_ela\_large\_r1} features since they are highly similar to the \texttt{deep\_ela\_large\_r10} features. From the four investigated \texttt{transoptas} variants, we include only \texttt{transoptas\_5DE} and \texttt{transoptas\_5PSO} since they are trained on algorithm portfolios consisting of completely different algorithm classes (DE and PSO), and it is of our interest to analyze whether similar representations are learned in this case. We remove the features from each feature group which have constant values for all problem instances.
Since the \texttt{tinytla} feature group contains a total of 5,050 features, which is a prohibitively high number for visualization, upon consultation with the \texttt{tinytla} creators, we reduced the number of features. In particular, we performed Principal Component Analysis (PCA) dimensionality reduction to reduce the features of each homology to 20 dimensions, which preserves 99\% of the explained variance. Please note that this transformation of the \texttt{tinytla} features is only done for visualization purposes for Figures~\ref{fig:correlation} and~\ref{fig:correlation_ela_subset}, and the features are used in their original form in the remaining analyses.

Along the colorbar, there are several large clusters composed entirely of \texttt{tinytla} features, a few clusters containing only \texttt{ela\_scaled} features, and a few clusters consisting of the two variants of \texttt{transoptas} features.
The \texttt{transoptas\_5DE} and \texttt{transoptas\_5PSO} are commonly clustered together, indicating that even though the features were learned from portfolios of different algorithm classes, they capture similar information. However, the clusters they produce have varying correlation levels, meaning that there are subgroups within these features that capture diverse information.

The \texttt{deep\_ela\_large\_r10} and \texttt{doe2vec} features occasionally produce smaller homogeneous clusters (with features coming from the same group), but are generally placed in diverse clusters.
Looking at the three largest clusters of \texttt{ela\_scaled} features, we can see that the biggest cluster is composed of the \texttt{disp} (dispersion) subgroup. The features within this cluster also have a high positive correlation with each other. The second largest cluster contains the features capturing the runtime costs from different subgroups, while the third largest one contains features from the \texttt{ela\_level} subgroup.

Figure~\ref{fig:correlation_ela_subset} focuses on the comparison of the previously investigated feature groups to the \texttt{ela\_scaled} features. Here, the \texttt{ela\_scaled} features are featured on the vertical axis, while the remaining feature groups are on the horizontal axis. The goal of this analysis is to gain some intuition about the information captured by the deep-learning based features, which are not interpretable, by relating them to some of the \texttt{ela\_scaled} features. From this figure, we can see that only a few of the \texttt{tinytla} features have strong correlations with the \texttt{ela\_scaled} features, while the majority of them are not correlated. However, here it is important to stress that these features have undergone a PCA dimensionality reduction to reduce the number of features for visualization purposes. Looking at the \texttt{doe2vec} features, we can see that most features are not correlated with the \texttt{ela\_scaled} features. Individual \texttt{doe2vec} features have strong correlations with the \texttt{ela\_scaled\_nbc} subgroup and all of the features in the \texttt{ela\_scaled\_disp} subgroup. On the other hand, for the \texttt{deep\_ela} we observe stronger correlations with different subgroups. The \texttt{transoptas} variants exhibit strong correlations with some specific features. In particular, a lot of the features are strongly correlated with the \texttt{ela\_scaled\_distr} subgroup, capturing properties of the distribution of objective function values, as well as the \texttt{ela\_scaled\_ic} subgroup, containing the information content features, and the \texttt{ela\_scaled\_nbc.nb\_fitness.cor} feature.

To further understand the complimentarity of different feature groups, our next experiment focuses on evaluating the consistency of similarity of problem pairs produced by different feature groups. The goal of this experiment is to identify feature groups where there is ``disagreement" about whether two problems are similar to each other or not. Pairs of features producing different similarities for the same pair of problems can be considered an indicator that these features capture diverging views of the problem instance and may be complimentary.

\begin{figure*}
    \centering
    \includegraphics[width=\linewidth]{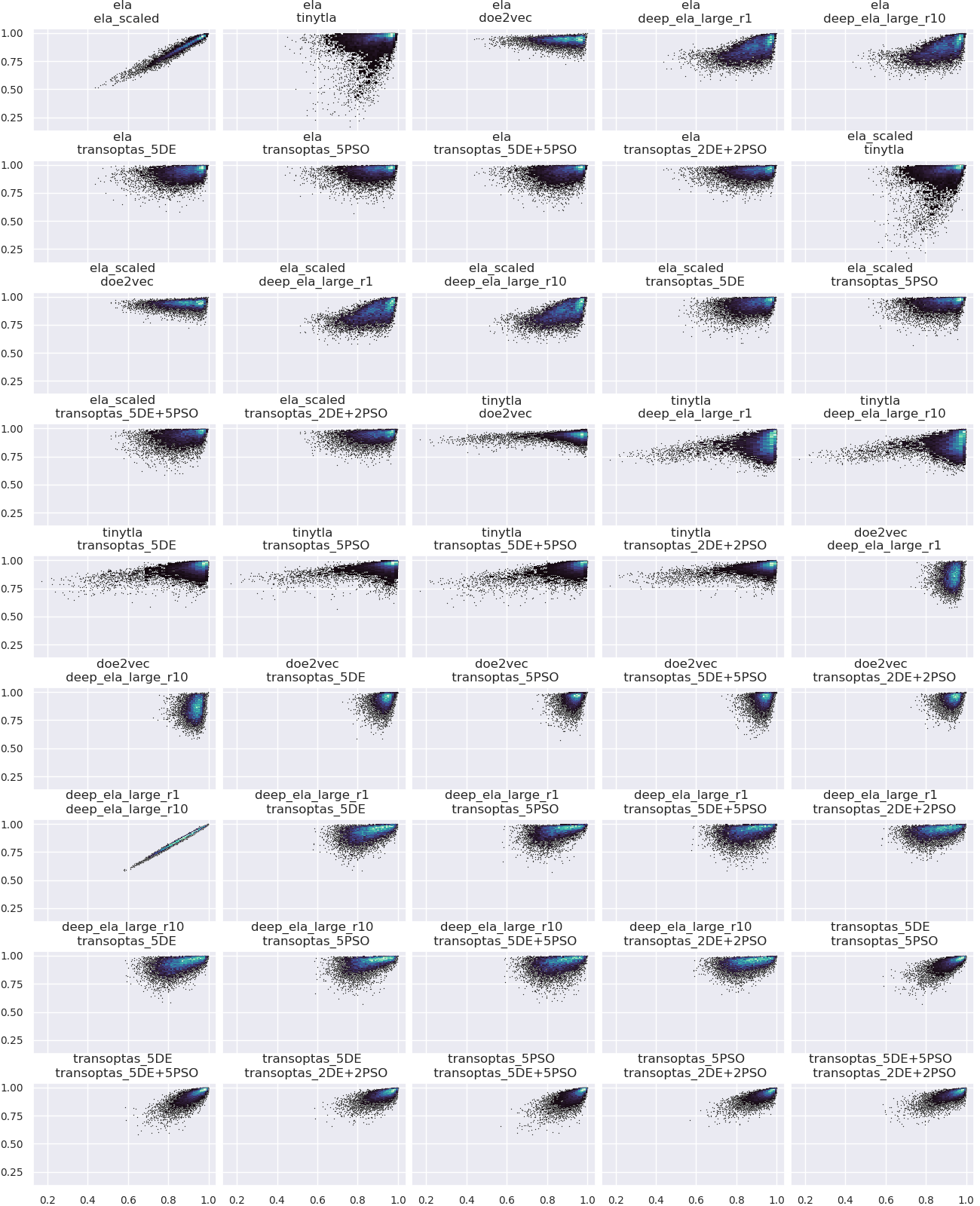}
    \caption{Cosine similarity of problem pairs, calculated with different feature sets. The first feature name in the title of each subplot refers to the horizontal axis, while the second feature name refers to the vertical axis.}
    \label{fig:problem_pair_similarity}
\end{figure*}

To this end, we randomly sample 10,000 pairs of problems and calculate the cosine similarities between their representations produced with each feature group. Before cosine similarity is calculated, constant features are removed and the remaining features are scaled to be in the range [0,1] using min-max scaling.
Figure~\ref{fig:problem_pair_similarity} shows the cosine similarity produced by each combination of features. The pair of problems are plotted in the form of a scatterplot, where the two coordinates denote the cosine similarity between these problems, obtained with two different feature representations.

The first feature name in the title of each subplot refers to the horizontal axis, while the second feature name refers to the vertical axis. For enhanced visibility, a histogram denoting the distribution of values is plotted on top of the scatterplot, where brighter values indicate higher values.
In the case of perfect consistency in the similarities produced by two feature groups, the subplot would resemble a diagonal bottom-up line. We can immediately see that this is the case for the \texttt{deep\_ela\_large\_r1} and \texttt{deep\_ela\_large\_r10} features, as well as the \texttt{ela} and \texttt{ela\_scaled} features. 
When problems are represented in terms of the \texttt{ela} and \texttt{ela\_scaled} features, the least similar problem pairs have similarities around 0.50. On the other hand, in terms of the \texttt{tinytla} features, the lowest similarities reach as low as 0.25.
Looking at the comparison of the \texttt{tinytla} features to other feature groups, we can see that there are some problem pairs where \texttt{tinytla} features produce similarities as low as 0.25, whereas other feature groups produce similarities above 0.80. This indicates that \texttt{tinytla} features capture information which is sometimes different than the other feature groups.

Focusing on the comparison of the \texttt{transoptas} features with other \texttt{transoptas} features, we can see that the shape of the distributions produced is quite similar. The cosine similarity of most of the problem pairs exceeds 0.80, with the least similar problem pairs having similarities as low as 0.70.

In summary, when comparing feature sets, we observe that in most cases there is some agreement between feature groups when problem instances are extremely similar. When one group assigns high similarity ($>$ 0.95), it usually means that the other feature groups will also assign high similarity. However, when pairs of instances are not as similar, one feature set might assign high similarity while another might assign low similarity, indicating a disagreement between feature sets on how similar the pairs of problem instances are.

\subsection{Algorithm Performance Alignment to Problem Landscape Features}
\label{subsec:results_feature_performance_alignment}
In order for a ML model to work, the input features should be able to capture the target variable.
To understand to what extent this is true for each of the feature types, in this section, we analyze the alignment of each feature type to algorithm performance. 

\begin{figure*}
    \centering
     \begin{subfigure}{0.32\textwidth}{\includegraphics[width=\textwidth]{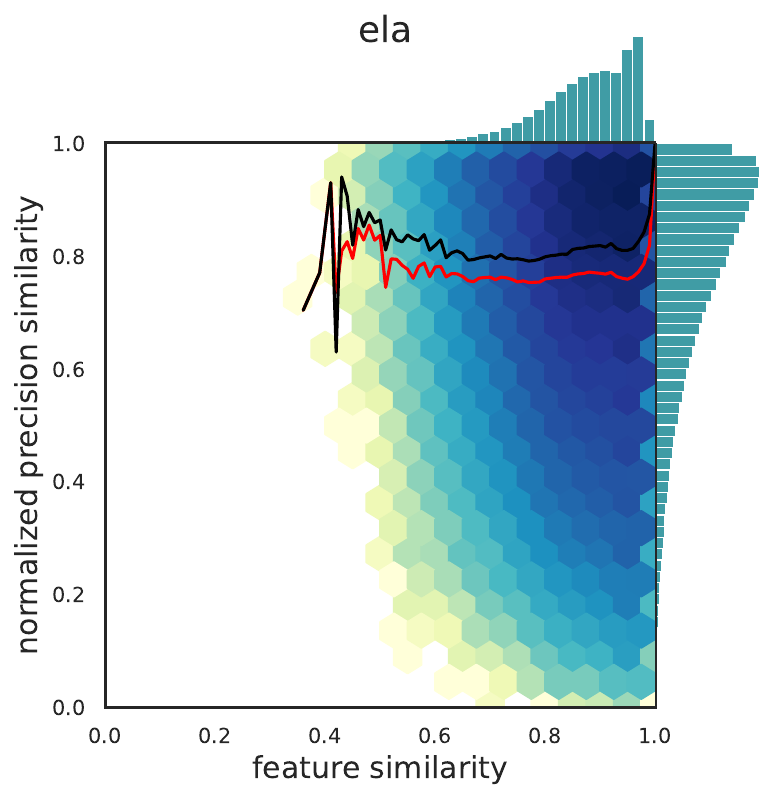}} 
    \end{subfigure}
\hfill
      \begin{subfigure}{0.32\textwidth}{\includegraphics[width=\textwidth]{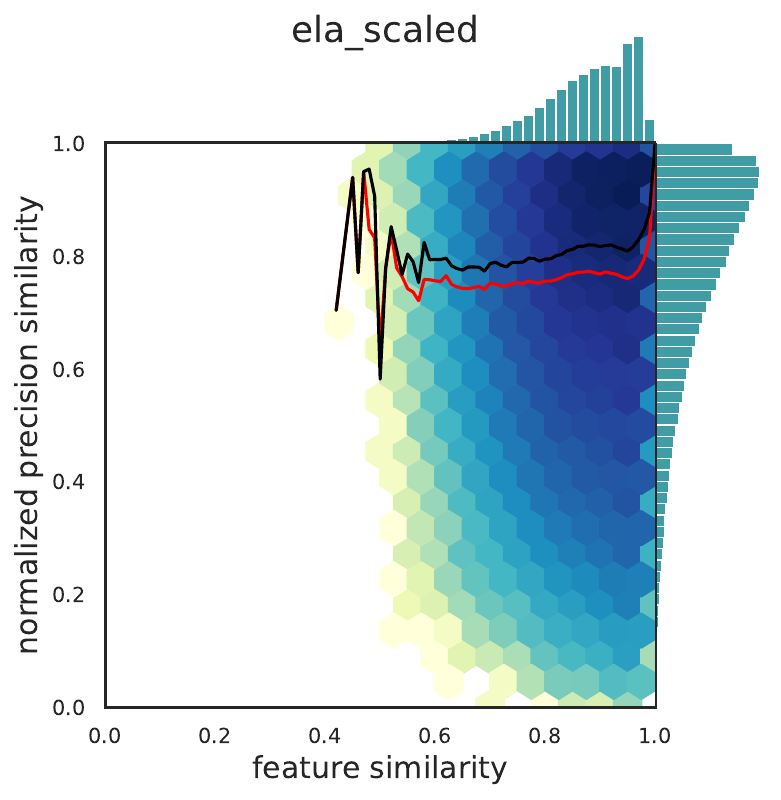}}
    \end{subfigure}
    \hfill
    \begin{subfigure}{0.32\textwidth}{\includegraphics[width=\textwidth]{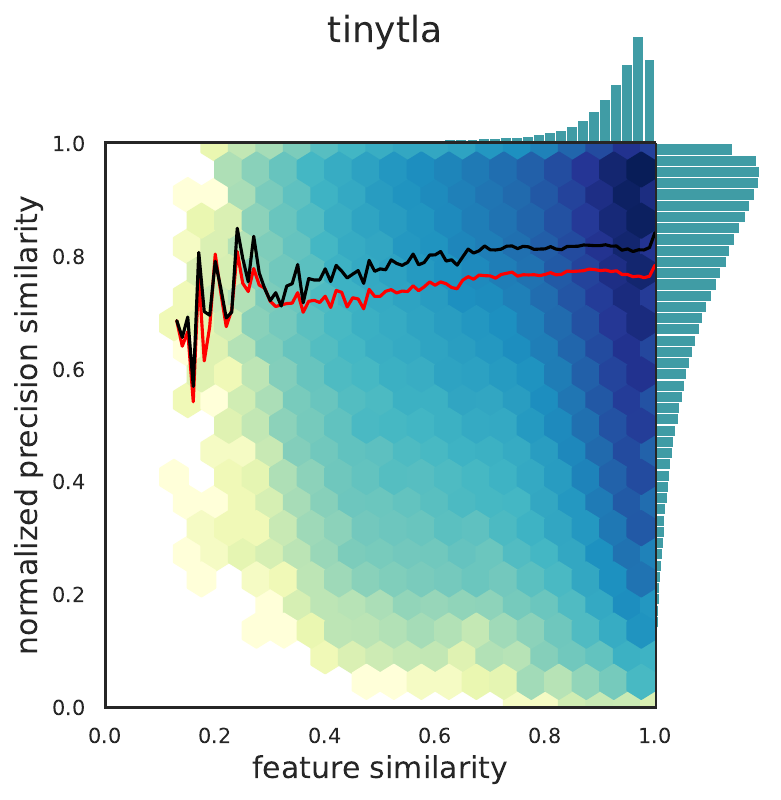}} 
    \end{subfigure}
\hfill
    \begin{subfigure}{0.32\textwidth}{\includegraphics[width=\textwidth]{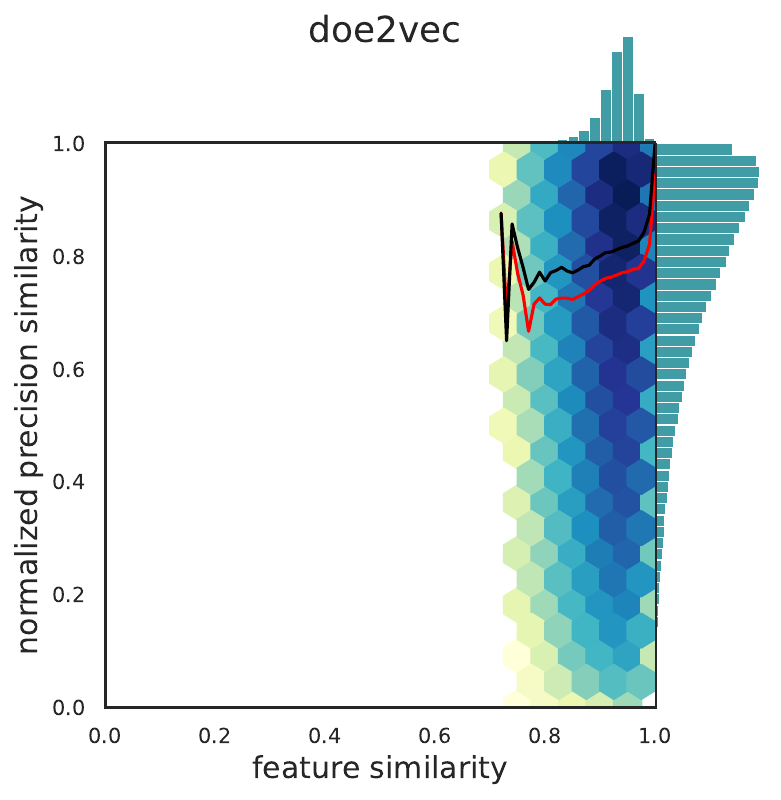}} 
    \end{subfigure}
    \hfill
      \begin{subfigure}{0.32\textwidth}{\includegraphics[width=\textwidth]{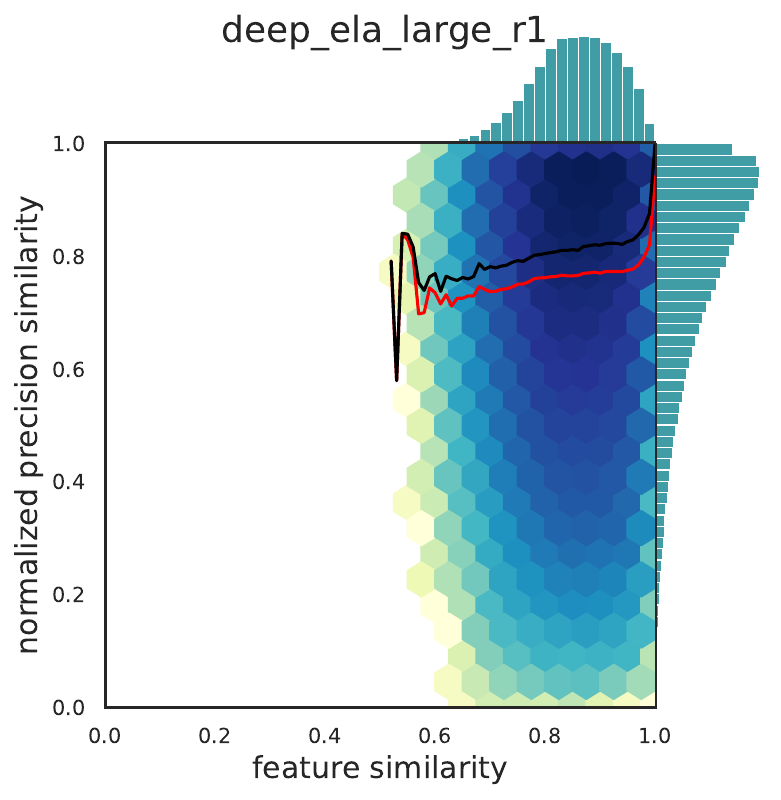}} 
    \end{subfigure}
\hfill
\begin{subfigure}{0.32\textwidth}{\includegraphics[width=\textwidth]{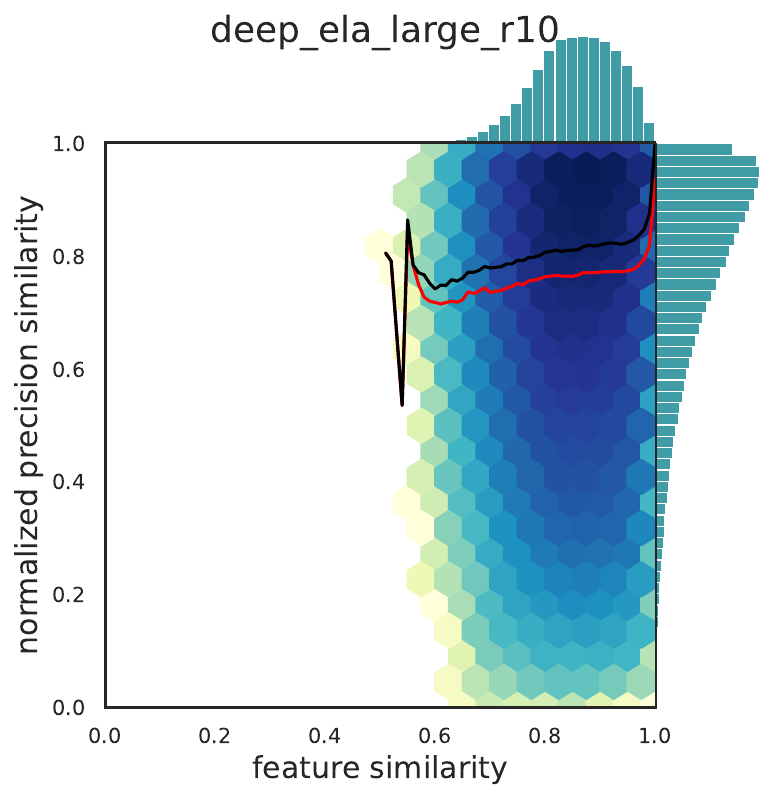}} 
    \end{subfigure}

\hfill
\begin{subfigure}{0.32\textwidth}{\includegraphics[width=\textwidth]{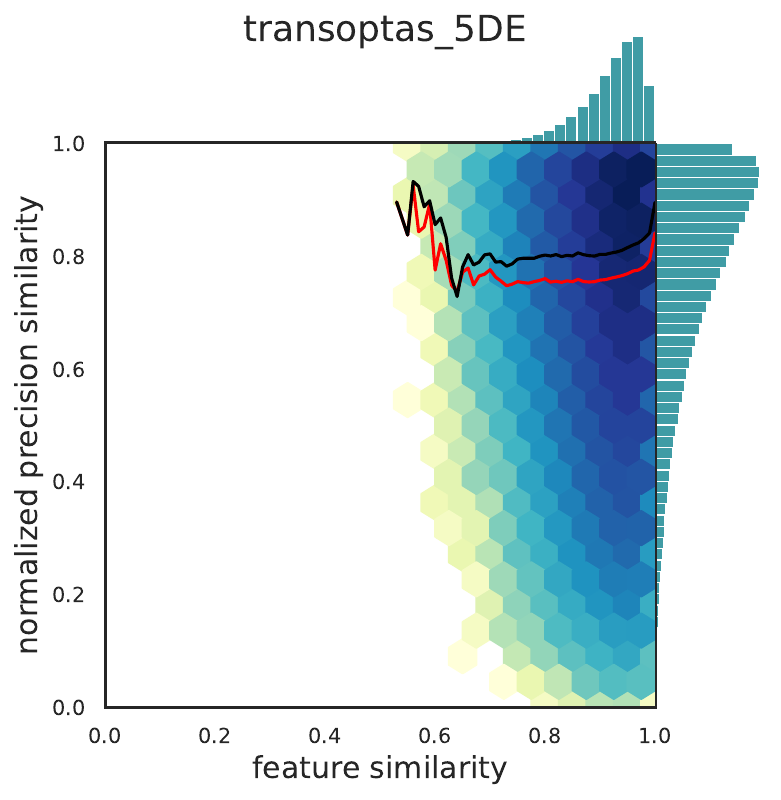}} 
    \end{subfigure}
\hfill
\begin{subfigure}{0.32\textwidth}{\includegraphics[width=\textwidth]{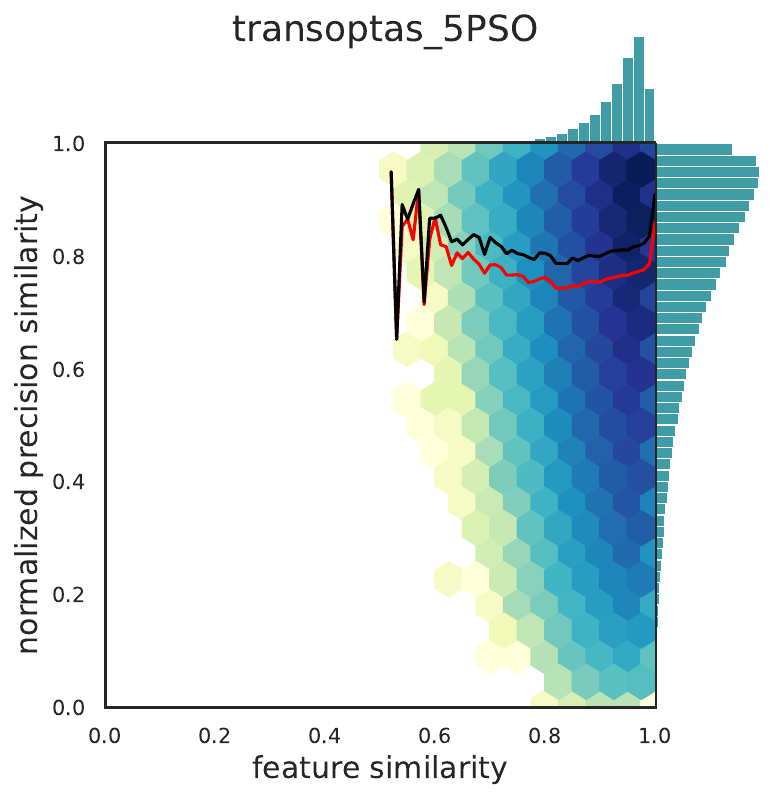}} 
    \end{subfigure}
\hfill
     \begin{subfigure}{0.32\textwidth}{\includegraphics[width=\textwidth]{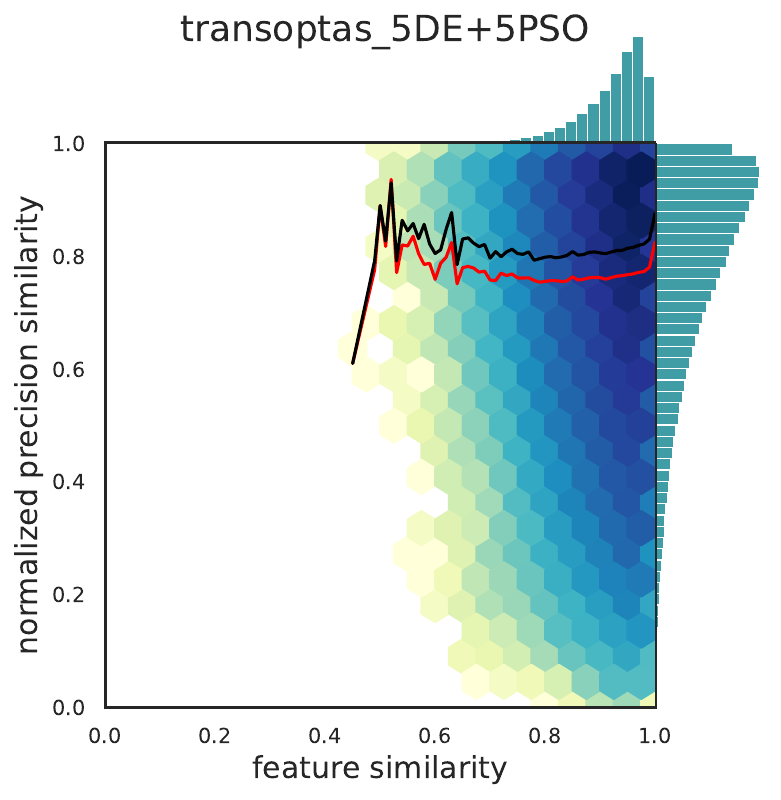}} 
    \end{subfigure}
    \hfill
     \begin{subfigure}{0.34\textwidth}{\includegraphics[width=\textwidth]{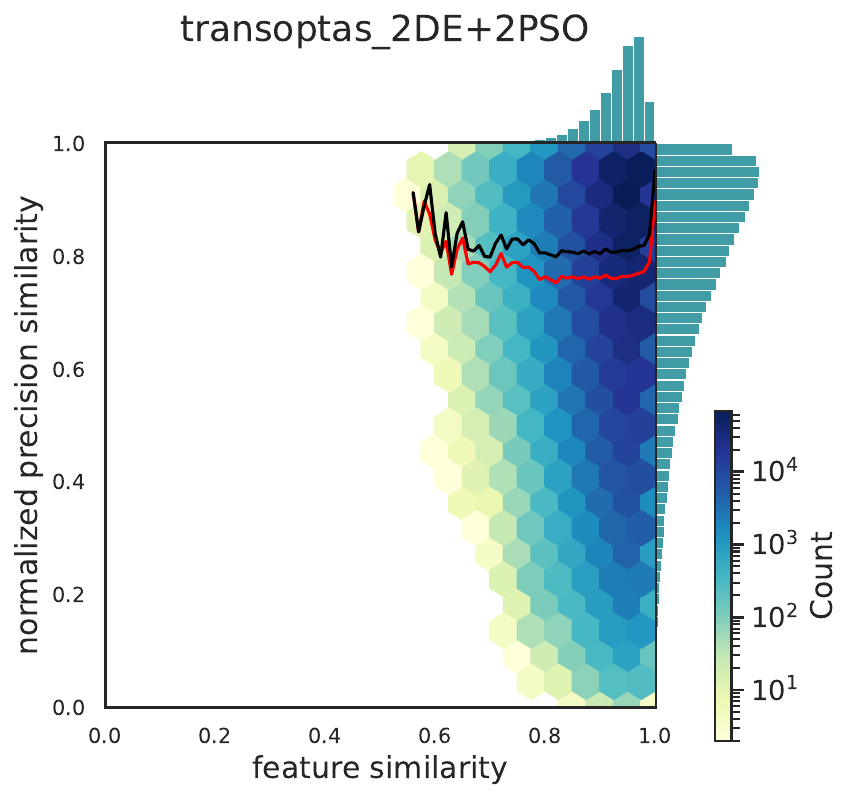}} 
    \end{subfigure}
    
    \caption{Distribution of feature and performance similarities for the 2DE+2PSO portfolio for 1,000,000 problem pairs}
    \label{fig:feature_performance_alignment}
\end{figure*}

As a first experiment, we investigate whether problems with similar feature representations exhibit similar algorithm performance. More specifically, we sample 1,000 of the affine problems and measure the cosine similarity of their feature representations and the cosine similarity of their performance vectors, containing the score of each of the algorithms in the algorithm portfolio. Figure~\ref{fig:feature_performance_alignment} shows the results for each feature type, for the 2DE+2PSO portfolio. We chose this portfolio for illustration purposes, since it exhibits the highest algorithm complementarity.
Each subplot shows the results for a different feature type. The horizontal axis shows the similarity in feature space, while the vertical axis shows the similarity in performance space. Both axis are accompanied by a histogram representation of the marginal distribution. The red line represents the mean value of the performance similarity for a given feature similarity (rounded to two decimal points), while the black line denotes the median.
 The main plot area displays hexagonal bins, where the color intensity of each hexagon indicates the number of data points within that bin. Darker hexagons represent areas with a higher concentration of data points. Please note that the number of points within each hexagon has undergone a logarithmic transformation before being mapped to its corresponding color, to prevent from hexagons having a low number of points being shown as empty.
In an ideal case, the feature similarity would be equal to the performance similarity, i.e., most of the points would be located on the diagonal from the top right to the bottom left corner. However, we can see that this is not the case for any feature type.
 
 From the marginal distributions, we can see that the feature similarities for all feature types are almost entirely above 0.60, with peaks in the distributions around 0.80-0.90. 
 The \texttt{ela} and \texttt{ela\_scaled} features have a median similarity of 0.89, the \texttt{tinytla}, \texttt{doe2vec}, and all of the \texttt{transoptas} features have median similarities around 0.94, while the \texttt{deep\_ela} features have median similarities around 0.85.
 Looking at the red and black lines, denoting the mean and median performance similarity for a given feature similarity, we can see that performance similarity is not usually increasing with the increase of feature similarity. We can only observe an increase when the values of the feature similarity exceed 0.98. In the areas with low feature similarity, we can also observe a zig-zag pattern appearing at the beginning of each line, which can be attributed to a low number of points with high variability in performance similarity located in this area.
 
 From this figure, we conclude that none of the feature groups exhibit a perfect alignment with algorithm performance. The fact that alignment between similarity in features and similarity in performance is only clearly observed when feature similarity is extremely high could indicate that an AS model could correctly infer performance if it has previously observed a similar or almost identical problem instance.

One weakness of the previous experiment is that when measuring the similarity of problem instances using all of the features within one feature group, there may be some features that capture algorithm performance, and others that do not. As a consequence, in the event when a majority of features within a single group are not capturing performance, the results of the previous experiment may be too pessimistic.
For that reason, we conduct another experiment which compares the correlation of algorithm performance with individual features within a feature group.

\begin{figure*}
      \centering
    \includegraphics[width=0.9\textwidth]{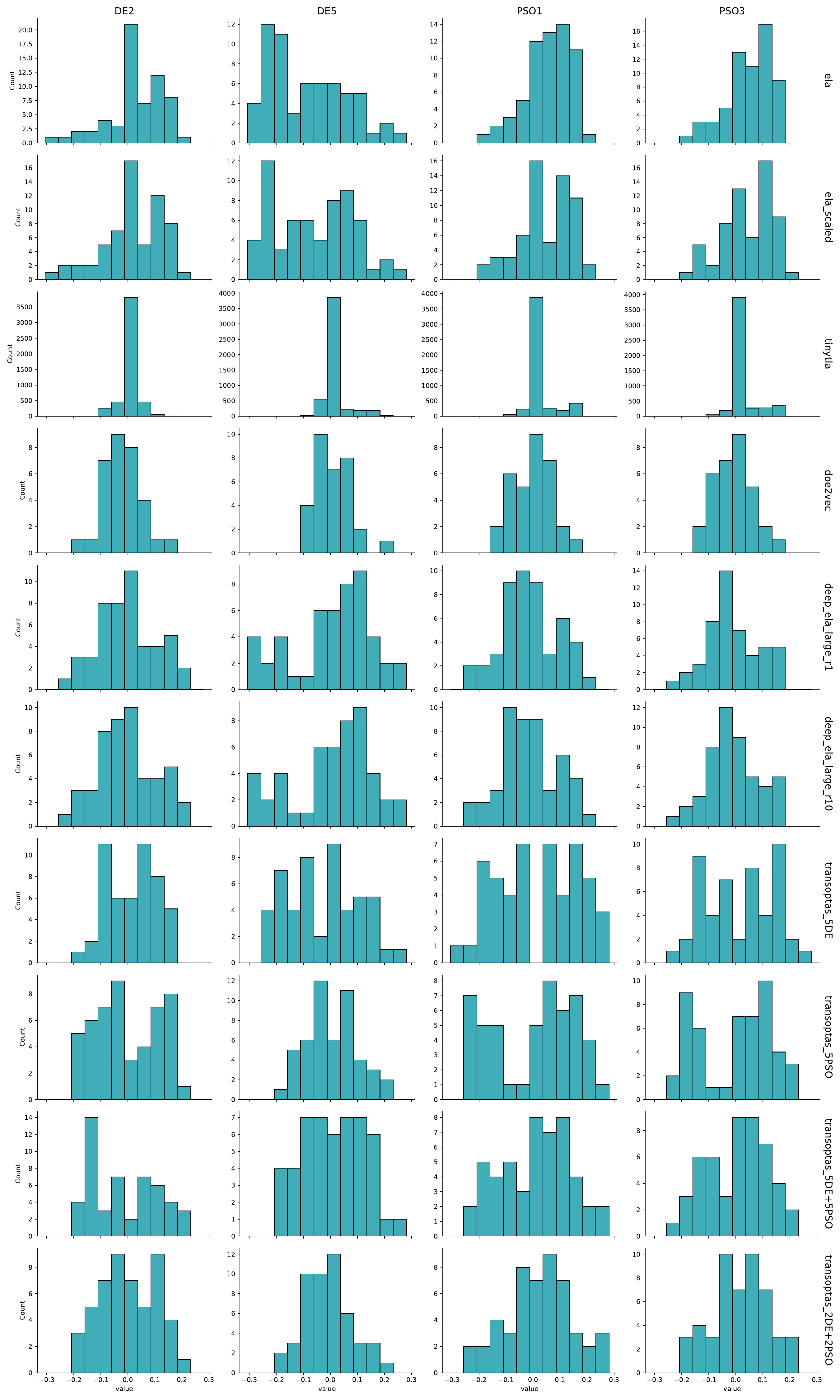}
    \caption{Distribution of Spearman correlation coefficient of individual features within feature groups with performance of individual algorithms within the 2DE+2PSO algorithm portfolio}
    \label{fig:single_feature_performance_correlation}
\end{figure*}

To this end, we calculate the Spearman correlation coefficient between each feature, and the performance (normalized precision) of individual algorithms. Figure~\ref{fig:single_feature_performance_correlation} shows the distribution of Spearman correlation coefficient for features within the feature groups. Each column denotes a single algorithm from the 2DE+2PSO portfolio, while each row denotes one of the different feature groups. In an ideal case, the distribution within each subplot should be bimodal with peaks along the leftmost or rightmost edge, meaning that most of the features should be correlated (positively or negatively) with algorithm performance. A peak at the center (zero) indicates that a lot of features within a feature group are not correlated with performance. In this case, we do not remove the constant features, since we want to consider how many of the features produced from each feature group are actually useful for capturing performance.
As a first observation, we can see that all correlations are between -0.30 and 0.30, indicating that none of the features individually have a very high correlation with algorithm performance.
We can see that the distributions of the \texttt{ela} and \texttt{ela\_scaled} features look relatively similar, which is unsurprising since a lot of the features are not sensitive to the scaling of objective function values and are therefore equivalent in both groups. 
Looking at the \texttt{tinytla} features, we can see that the disribution has a large peak at zero. This is due to the fact that a large portion of the \texttt{tinytla} have constant values, and their Spearman correlation is therefore set to zero. The \texttt{doe2vec} features are are also centered at the zero point, with the most highly correlated feature being around -0.20 and 0.20. The \texttt{deep\_ela\_large\_r1} and \texttt{deep\_ela\_large\_r10} have nearly identical distributions, with a slightly higher spread than the \texttt{doe2vec} features. Interestingly, for the DE5 algorithm, there is a larger number of \texttt{deep\_ela} features with a negative correlation.
Considering the four \texttt{transoptas} variants, we can observe that in several cases, the distributions are bimodal and not centered at zero. This is the case for the \texttt{transoptas\_5DE} and \texttt{transoptas\_5PSO} features for most of the algorithms.

In conclusion, this analysis shows that none of the investigated feature groups are perfectly capturing algorithm performance, and all of the feature groups have individual features which are not at all correlated with algorithm performance.

\subsection{Algorithm Selection Results}
\label{subsec:results_AS}
\begin{figure*}
      \centering
    \includegraphics[width=\textwidth]{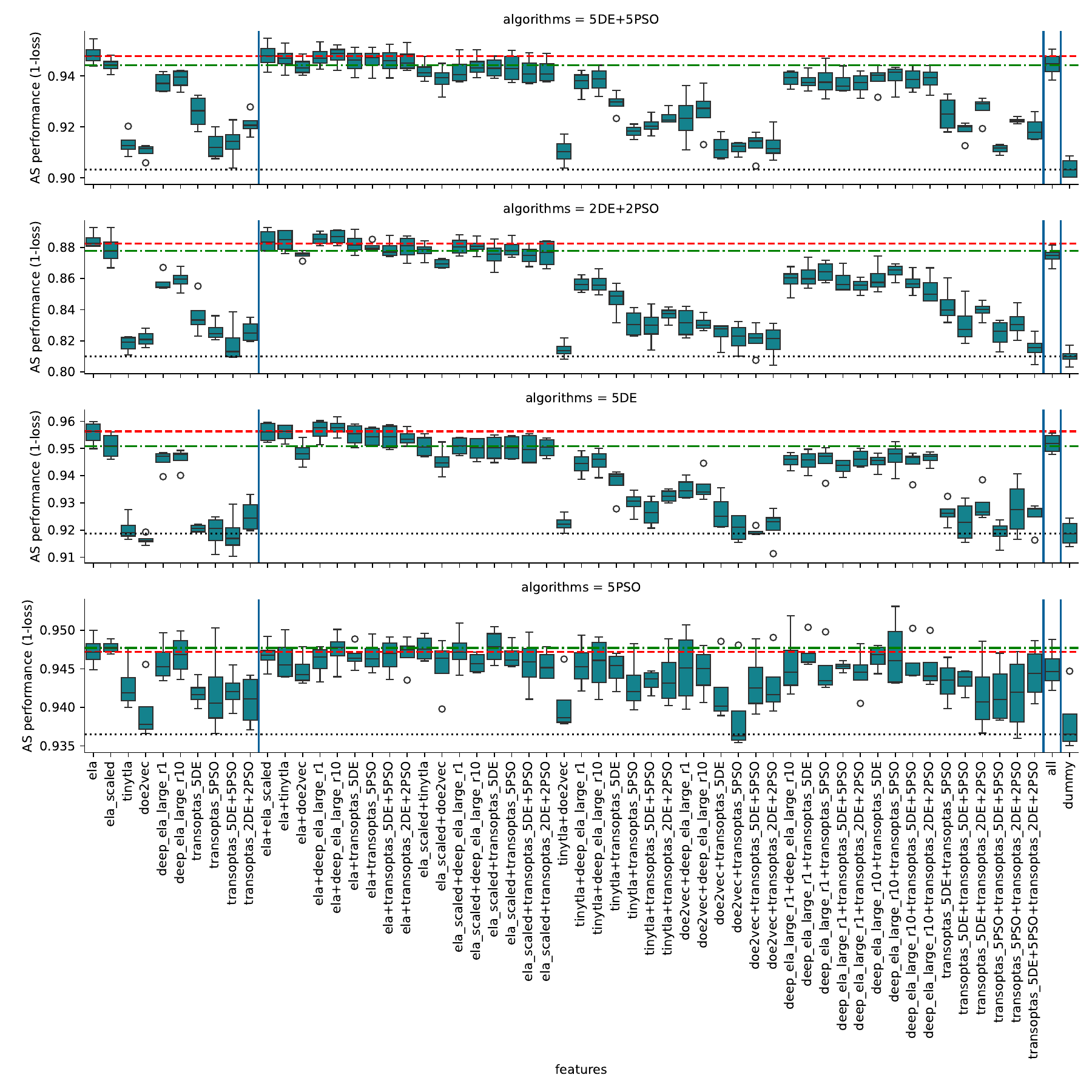}
    \caption{AS performance obtained for the different feature groups and algorithm portfolios in the instance split evaluation}
    \label{fig:AS_I_split}
\end{figure*}

\begin{figure*}
      \centering
    \includegraphics[width=\textwidth]{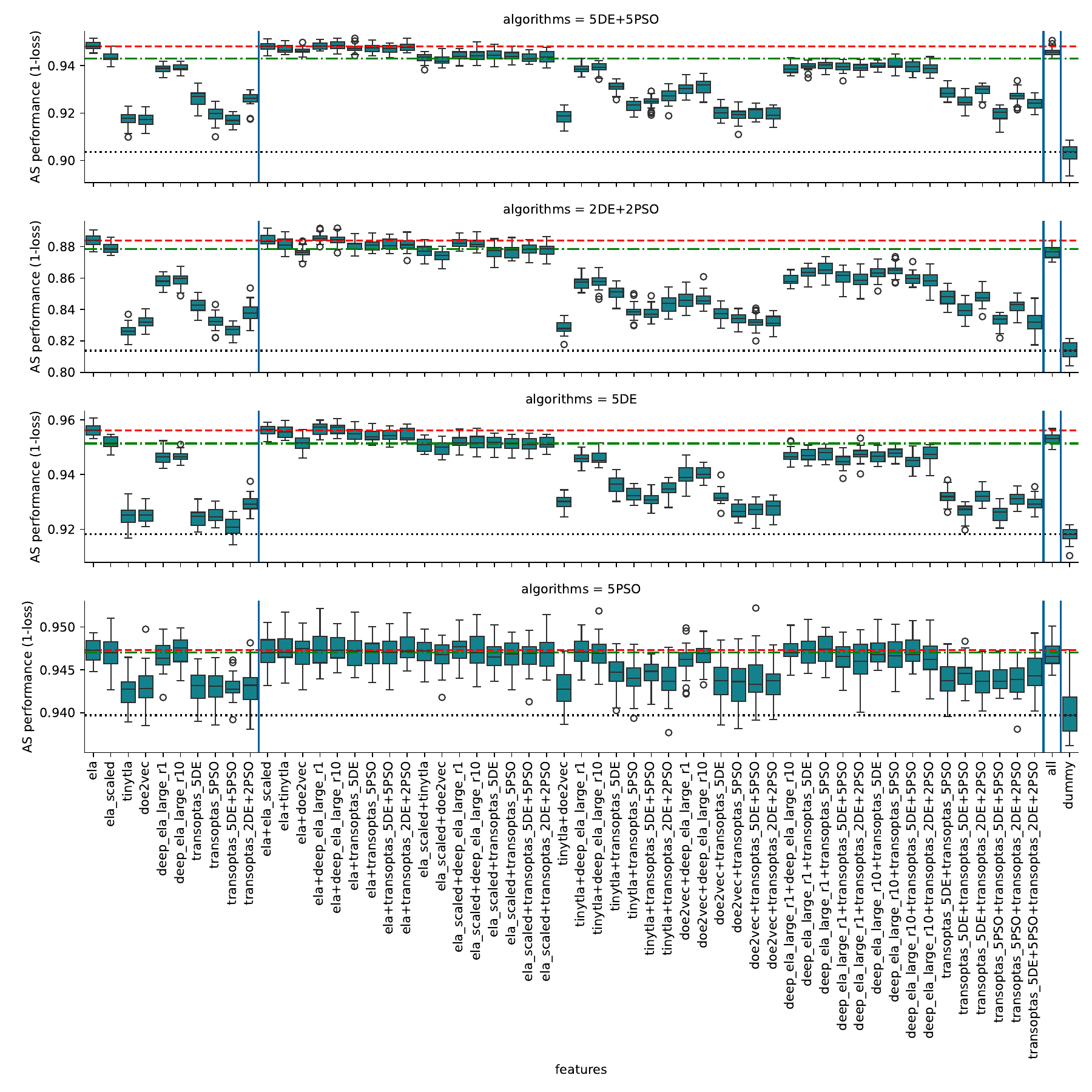}
      \caption{AS performance obtained for the different feature groups and algorithm portfolios in the random split evaluation}
    \label{fig:AS_R_split}
\end{figure*}

\begin{figure*}
      \centering
    \includegraphics[width=\textwidth]{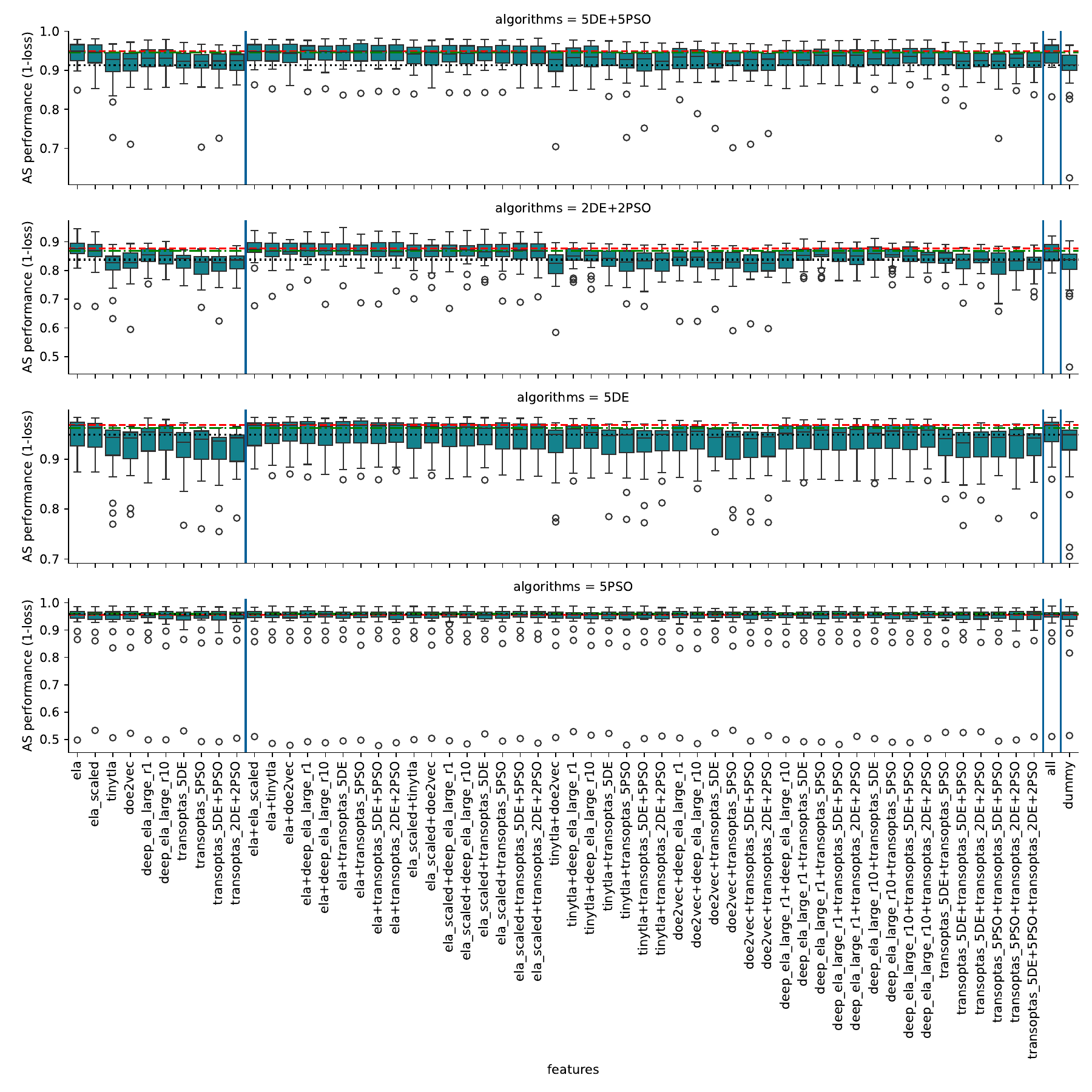}
      \caption{AS performance obtained for the different feature groups and algorithm portfolios in the problem combination split evaluation}
    \label{fig:AS_PC_split}
\end{figure*}

\begin{figure*}
      \centering
    \includegraphics[width=\textwidth]{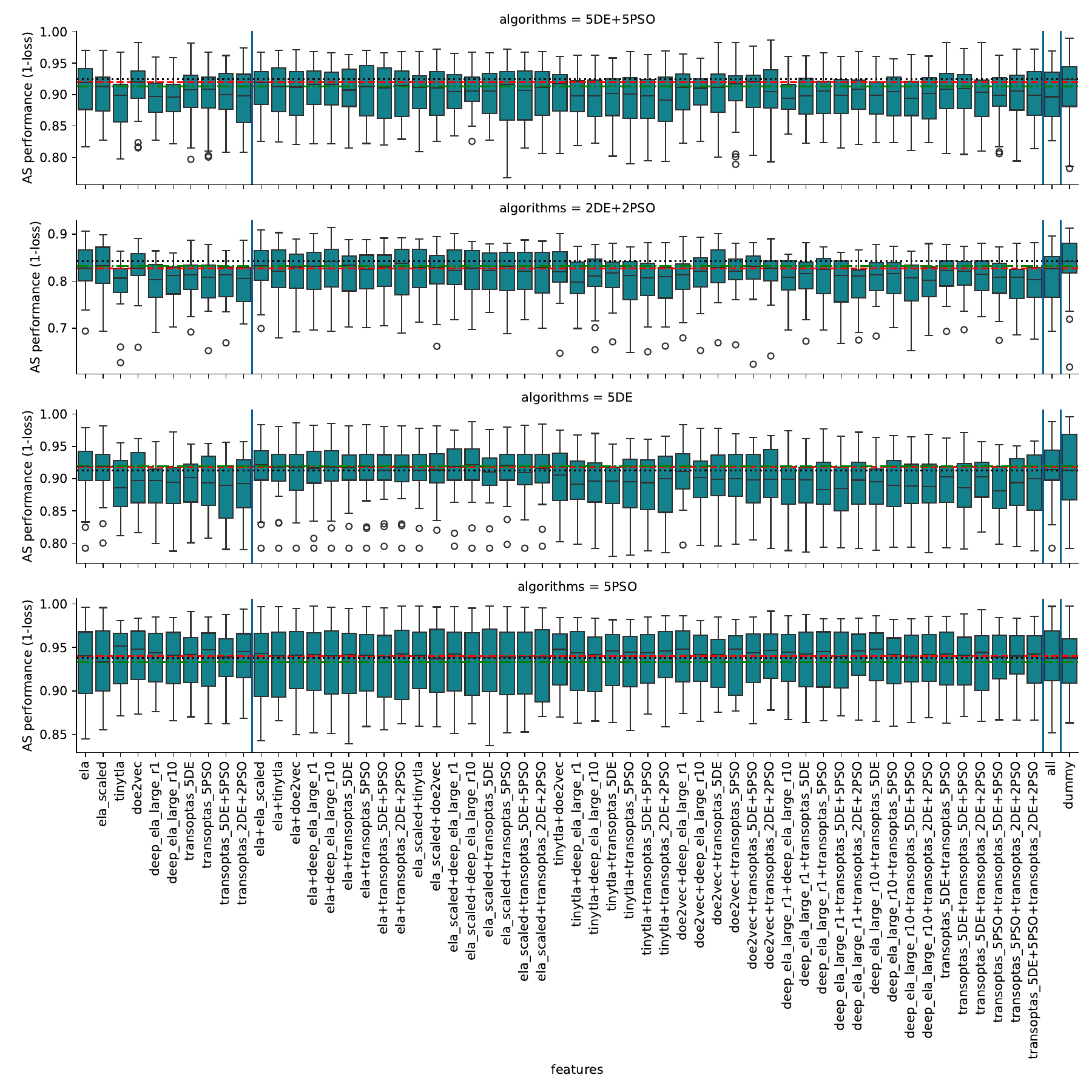}
      \caption{AS performance obtained for the different feature groups and algorithm portfolios in the problem split evaluation}
    \label{fig:AS_P_split}
\end{figure*}

In this subsection, we present the results from the AS in terms of the AS performance metric defined in Equation~\ref{eq:AS_metric}. Figures~\ref{fig:AS_I_split},~\ref{fig:AS_R_split},~\ref{fig:AS_PC_split} and~\ref{fig:AS_P_split} present the results from the four different evaluation settings: Instance split, Random split, Problem Combination split, and Problem split.
Each figure is divided into four subfigures, containing the result for a different algorithm portfolio. The horizontal axis contains the names of the features used to train the model, while the vertical axis contains the value of the AS performance metric defined in Equation~\ref{eq:AS_metric}, indicating the performance of the model using the particular set of features for the AS task. Please recall that the AS metric is supposed to be maximized.
In each subplot, there are three horizontal and vertical lines added with the purpose of increasing readability.
The three vertical lines are meant to separate the features into groups based on the number of feature groups being combined. In the first group, each feature group is used individually. In the second group, pairs of feature groups are being combined. The third group consists of all of the feature groups being combined together, while the fourth group is the baseline dummy model.
The red dashed horizontal line denotes the median AS performance of the original \texttt{ela} features, the green dash-dotted line denotes the median AS performance of the \texttt{ela\_scaled} features, while the black dotted line denotes the median AS performance of the baseline dummy model. These lines are meant to provide an easier comparison of the results obtained by different feature groups to the results of the dummy model and the \texttt{ela} features.

Taking into consideration the results from the instance split evaluation in Figure~\ref{fig:AS_I_split}, we can see that all of the feature-based models generally outperform the dummy model. 
The difference between the median AS performance of the \texttt{ela} model and the dummy model amounts to 0.041, 0.067, 0.032, and 0.011 for the 5DE+5PSO, 2DE+2PSO, 5DE and 5PSO algorithm portfolios, respectively.

The \texttt{ela\_scaled} features underperform the \texttt{ela} features except for the 5PSO portfolio, where they outperform \texttt{ela} by a small margin. This means that scaling the objective function values before feature calculation is generally not beneficial in this case. Comparing the results in the first horizontal group, where individual feature groups are used, we can see that the \texttt{ela} and \texttt{ela\_scaled} features are providing the best results, with the two variants of the \texttt{deep\_ela} features being the second ranked, the \texttt{transoptas} variants ranking third, followed by \texttt{tinytla} and \texttt{doe2vec}.

Looking into the second horizontal group, where pairs of feature groups are combined together, we observe that most of the feature pairs still do not manage to outperform the individual use of the \texttt{ela} features. In some rare cases, we do observe a very minor improvement in performance. This is the case when combining the \texttt{ela} with the \texttt{deep\_ela} variants for the 2DE+2PSO portfolio and 5DE portfolio, although the difference is negligible.

Using all of the features together results in no improvement over the individual use of the \texttt{ela} features.

Looking at Figure~\ref{fig:AS_R_split}, showing the results for the random split, we observe a similar pattern of the feature-based models outperforming the dummy model, and the individual use of the \texttt{ela} features providing the best results overall with negligible improvements contributed to the other feature groups. The \texttt{ela} model outperforms the dummy model by 0.039, 0.064, 0.033, and 0.007 for the 5DE+5PSO, 2DE+2PSO, 5DE and 5PSO algorithm portfolios, respectively.

Focusing on Figure~\ref{fig:AS_PC_split}, presenting the results from the problem combination split evaluation, where all problems derived from one parent problem class are used for testing, we can see that there is a drop in the improvement provided by the feature-based models compared to the dummy. 
In this case, difference between the median AS performance of the \texttt{ela} model and the dummy model corresponds to 0.032, 0.030, 0.013, and 0.001 for the 5DE+5PSO, 2DE+2PSO, 5DE and 5PSO algorithm portfolios, respectively. For the 5PSO algorithm portfolio, the feature-based models no longer outperform the dummy, or the difference in their performance is close to zero. We can also observe that there are folds for which the performance of all models is drastically lower than the others, which is likely an indicator that there is one problem class which when combined with remaining 23 problem classes, produces problems which are outside of the distribution of the other problems generated by combining the 23 problem classes.

Finally, Figure~\ref{fig:AS_P_split} presents the results from the most challenging problem split evaluation scenario, where all problems derived from a pair of problem classes are used for testing.
In this case, we can see that the AS is no longer working, with all models having essentially equivalent or inferior performance to the dummy model. The difference between the median AS performance of the \texttt{ela} model and the dummy model corrensponds to -0.011, -0.009, 0.006 and -0.004 for the 5DE+5PSO, 2DE+2PSO, 5DE and 5PSO algorithm portfolios, respectively.

Overall, the results indicate that while all feature groups outperform the baseline model and are useful for AS in the less challenging settings (instance and random split), where the testing set has similar instances as the training set. However, all of the features fail in the most challenging evaluation setting, where none of the feature groups outperform a simple baseline model. Additionally, even though the presented results are achieved with an algorithm execution budget of 100 iterations, our conducted experiments with different budgets yielded similar \changed{results and conclusions, which can be found in the Supplementary Materials.}
 The explored feature groups do not provide a substantial improvement over using just the \texttt{ela} features, although some of them may provide similar results at a lower feature calculation cost (assuming the deep learning model is already trained, computing the features is fast since the samples are only passed through the model). A statistical analysis of the obtained results can be found in the Supplementary Materials. 
\section{Discussion}
\label{section_discussion}
The results of this work show that while new representations for optimization problems have been developed, moving beyond the traditional ELA features, the challenge of generalizing to new, unseen problems persists. Meta-models work well on functions that are the same or very similar to the training data, but they still fall short of the main goal -- generalizing to entirely new functions.

Considering the fact that \texttt{doe2vec}, \texttt{deepela} and \texttt{transoptas} features are trained using different training strategies, and the \texttt{tinytla} features use a completely different approach to feature calculation, the obtained results may be an indicator that the features themselves are not the cause of the observed lack of generalizability to unseen data. On the other hand, our experiments show that the similarities between problem instances is quite high, regardless of the feature representation used.

This raises the question of what exactly is preventing the AS model from transferring the knowledge to new data. Our analysis of performance alignment to problem landscape features indicated that alignment between similarity in features and similarity in performance is only clearly observed when feature similarity is extremely high. This could explain the poor generalizability of the meta-model, since the AS model can reliably infer performance if it has previously observed a similar or almost identical problem instance. Other potential causes to be further investigated include:

\textit{i)} Is a 50$d$ Latin Hypercube sample of the problem sufficient to capture algorithm performance? Topics to be investigated here from an experimental setup point of view, include both the size of the sample and the sampling technique. Although the sensitivity of ELA features to the sample size and sampling method has been investigated in several works~\citep{urban_ela_not_invariant_sampling, urban_ela_not_invariant, ryan_sample_size}, to the best of our knowledge, this topic has not been investigated from an AS generalizability point of view. Since most of the AS works use the instance split evaluation, the sample size of 50$d$ has been accepted as the general recommendation. It could be the case that this sample size is sufficient for the easier instance split evaluation, but not for the more challenging evaluation. However, a larger sample size of 1000$d$ for calculating the ela features has been used in~\citep{gasper_affine_generalization}, and in this case, the AS model still did not work in the more challenging problem split evaluation.

\textit{ii)} All of the investigated feature groups based on deep learning models (\texttt{doe2vec}, \texttt{deepela} and \texttt{transoptas}) leverage a random function generator for creating the training data. The functions generated in this way have been shown to cover a much different area of the problem space compared to the affine functions~\citep{gina_cross_benchmark_generalization, gina_ela_generalization_scaling}. This discrepancy in the training data may be an additional factor influencing the AS generalizability. The improvement of deep learning models for feature extraction necessitates a further effort in developing large and diverse training benchmarks which provide a broad coverage of the problem space.

\textit{iii)} One possible explanation for the lack of generalizability is the design of the BBOB benchmark and its current use in the algorithm selection community. While BBOB problems are intended to be diverse, instances of a specific problem are often quite similar to each other. With that in mind, it is possible that no benchmark currently exists that would allow for a fair comparison of features, as one LIO approach might be trivially solved, while LPO is extremely challenging. This could potentially mean that, with existing benchmarks, it is not possible to measure generalizability properly.

The previously provided points are directions towards understanding why this experimental setting is failing. However, the overall results indicate that a more radical change is required in the way AS is approached for single-objective continuous optimization, especially in the way it is evaluated.

\changed{Finally, we would like to acknowledge a limitation of our experimental evaluation, which is the choice of benchmark suite. Namely, we use the affine problems generated by combining problems from the BBOB~\citep{bbob} benchmark. While we consider the affine problems to provide an improvement over the BBOB benchmark in terms of the quantity and diversity of the generated problems, we would like to point out that they are likely not sufficient to accurately model the vast majority of real-world problems.
In fact, several studies indicate that the generated affine problems are not too different from the base BBOB problems used for their construction. This has been observed by investigating the ELA representations of BBOB and affine problems, where it is shown that problems from both benchmarks have feature values in similar ranges~\citep{affine_training_size_selection, gina_cross_benchmark_generalization} and they cover the same area of the problem landscape when visualized in two dimensions~\citep{affine_training_size_selection, gina_cross_benchmark_generalization}.
}

\changed{
Unfortunately, to the best our knowledge, a sufficiently large benchmark of real-world problems is currently unavailable. Having all of this in mind, we still opted for using the affine combinations as they offer the creation of additional problems in a controlled setup, which are suitable for use in machine learning settings and evaluating the generalization of AS models. We do acknowledge the impact the training data has on our evaluation and conclusions, and would like to stress the importance of evaluating AS models on real-world problems, once they become available. 
}

\section{Conclusion}
\label{section_conclusion}
Our study highlights the limitations of current feature-based AS models in generalizing to out-of-distribution problems within single-objective continuous optimization. Despite the advancements in landscape analysis and deep learning-based features, when faced with unseen problems, these models fail to surpass the performance of a simple baseline model. This underscores the need for further research in the rethinking and development of more robust AS models capable of effectively handling novel problem instances.
A potentially promising research direction is the exploration of trajectory features which are calculated based on the candidate solutions which an algorithm is exploring during its execution, with examples such as~\citep{opt2vec, probing_trajectories, cenikj2023dynamorep}. Additionally, in this work, we evaluate the features for fixed-budget AS. As future work, the generalizability of these features can be evaluated in the fixed-target scenario also. \changed{Finally, we would like to once again stress the importance of evaluating AS models on real world problems. This is not possible in our case due to the limitated number of real world problems available in current benchmarks, but is an important next step to be considered once this limitation is overcome.}

\section*{Acknowledgments}
Funding in direct support of this work: Slovenian Research Agency: research  core  funding  No. P2-0098, young researcher grants No. PR-12393 to GC and No. PR-11263 to GP, project GC-0001 and project J2-4660.

\section*{Declaration of generative AI and AI-assisted technologies in the writing process}

During the preparation of this work the authors used ChatGPT in order to correct grammar and spelling mistakes. After using this tool/service, the authors reviewed and edited the content as needed and take full responsibility for the content of the publication.






 \bibliography{main}
\end{document}